\documentclass{article}

\usepackage{microtype}
\usepackage{graphicx}
\usepackage{subfigure}
\usepackage{booktabs} \usepackage{todonotes}
\usepackage{hyperref}

\usepackage[accepted]{icml2020}

\usepackage{amssymb,amsmath,amsthm}
\newtheorem{theorem}{Theorem}
\newtheorem{definition}{Definition}
\newtheorem{prop}{Proposition}
\newtheorem{lemma}{Lemma}
\newtheorem{corollary}{Corolary}

\newcommand\eg{\emph{e.g.}}
\newcommand\Real{\mathbb{R}}
\newcommand\st{~~~~\text{s.t.}~~~~}

\def\Xcal{{\mathcal X}}

\def\Hcal{{\mathcal H}}
\def\Fcal{{\mathcal F}}
\def\Ecal{{\mathcal E}}

\def\R{{\mathbb R}}

\def\Vcal{{\mathcal V}}
\def\Ecal{{\mathcal E}}
\def\Pcal{{\mathcal P}}
\def\Wcal{{\mathcal W}}
\def\Ncal{{\mathcal N}}
\def\Mcal{{\mathcal M}}

\icmltitlerunning{Convolutional Kernel Networks for Graph-Structured Data}

\begin{document}

\twocolumn[
\icmltitle{Convolutional Kernel Networks for Graph-Structured Data}

\begin{icmlauthorlist}
\icmlauthor{Dexiong Chen}{Inria}
\icmlauthor{Laurent Jacob}{CNRS}
\icmlauthor{Julien Mairal}{Inria}
\end{icmlauthorlist}

\icmlaffiliation{Inria}{Univ. Grenoble Alpes, Inria, CNRS, Grenoble INP, LJK, 38000 Grenoble, France}
\icmlaffiliation{CNRS}{Univ. Lyon, Universit\'e Lyon 1, CNRS, Laboratoire de Biom\'etrie et Biologie Evolutive UMR 5558, 69000 Lyon, France}

\icmlcorrespondingauthor{Dexiong Chen, Julien Mairal}{firstname.lastname@inria.fr}
\icmlcorrespondingauthor{Laurent Jacob}{laurent.jacob@univ-lyon1.fr}

\icmlkeywords{Graph Kernels, Kernel Methods, Deep Kernel Machines}

\vskip 0.3in
]

\printAffiliationsAndNotice{} 
\begin{abstract}
We introduce a family of multilayer graph kernels and establish
new links between graph convolutional neural networks and kernel methods. Our
approach generalizes convolutional kernel networks to graph-structured data, by
representing graphs as a sequence of kernel feature maps, where each node carries
information about local graph substructures. 
On the one hand, the kernel
point of view offers an unsupervised, expressive, and easy-to-regularize data
representation, which is useful when limited samples are
available. On the other hand, our model can also be trained end-to-end on large-scale data, leading
to new types of graph convolutional neural networks.  We show
that our method achieves competitive performance on several graph
classification benchmarks, while offering simple model interpretation.
Our code is freely available at \url{https://github.com/claying/GCKN}. \end{abstract}

\section{Introduction}
\label{sec:introduction}
Graph kernels are classical tools for representing graph-structured
data~\citep[see][for a
survey]{kriege2019survey}. Most successful examples represent graphs
as very-high-dimensional feature vectors that enumerate and count
occurences of local graph sub-structures. In order to perform well, a
graph kernel should be as expressive as possible, \emph{i.e.}, able to
distinguish graphs with different topological properties
\citep{kriege2018property}, while admitting polynomial-time algorithms
for its evaluation. Common sub-structures include
walks~\cite{gartner2003graph}, shortest
paths~\cite{borgwardt2005protein},
subtrees~\cite{shervashidze2011weisfeiler}, or
graphlets~\cite{shervashidze2009efficient}.

Graph kernels have shown to be expressive enough to yield good
empirical results, but decouple data representation and model
learning.
In order to obtain task-adaptive representations,
another line of research based on neural networks has been developed
recently~\cite{niepert2016learning,kipf2016semi,xu2018how,verma2018feastnet}.
The resulting tools, called graph neural networks (GNNs), are
conceptually similar to convolutional neural networks (CNNs) for
images; they provide graph-structured multilayer models, where each
layer operates on the previous layer by aggregating local neighbor
information.  Even though harder to regularize than kernel
methods, these models are trained end-to-end and are able to extract
features adapted to a specific task.  In a recent work,~\citet{xu2018how} have shown that
the class of GNNs based on neighborhood aggregation is at most as
powerful as the Weisfeiler-Lehman (WL) graph isomorphism test, on
which the WL kernel is
based~\cite{shervashidze2011weisfeiler}, and other types of network
architectures than simple neighborhood aggregation are needed for more
powerful features.

Since GNNs and kernel methods seem to benefit from different characteristics,
several links have been drawn between both worlds in the context of
graph modeling.  For instance, \citet{lei2017deriving} introduce a
class of GNNs whose output lives in the reproducing kernel
Hilbert space (RKHS) of a WL kernel. In this line of research, the kernel framework is
essentially used to design the architecture of the GNN since
the final model is trained as a classical neural network. This is also
the approach used by \citet{zhang2018end} and
\citet{morris2019weisfeiler}.  By contrast, \citet{du2019graph}
adopt an opposite strategy and leverage a GNN architecture to design new graph
kernels,
which are equivalent to infinitely-wide GNNs
initialized with random weights and trained with gradient descent.
Other attempts to merge neural networks and graph kernels involve
using the metric induced by graph kernels to initialize a
GNN~\cite{navarin2018pre}, or using graph kernels to obtain continuous
embeddings that are plugged to neural
networks~\cite{nikolentzos2018kernel}.

In this paper, we go a step further in bridging graph neural networks
and kernel methods by proposing an explicit multilayer kernel
representation, which can be used either as a traditional kernel
method, or trained end-to-end as a GNN when enough labeled data are
available. The multilayer
construction allows to compute a series of maps which account for
local sub-structures (``receptive fields'') of increasing size. The
graph representation is obtained by pooling the final representations
of its nodes. The resulting kernel extends to graph-structured data
the concept of convolutional kernel networks (CKNs), which was
originally designed for images and sequences~\cite{mairal2016end,chen2019bio}.
As our representation of nodes is built by iteratively aggregating
representations of their outgoing paths, our model can also be seen as
a multilayer extension of path kernels. Relying on paths rather than
neighbors for the aggregation step makes our approach more expressive
than the GNNs considered in~\citet{xu2018how}, which implicitly rely
on walks and whose power cannot exceed the Weisfeiler-Lehman (WL)
graph isomorphism test. Even with medium/small path lengths (which leads to reasonable
computational complexity in practice), we show that the resulting
representation outperforms walk or WL kernels.

Our model called graph convolutional kernel network (GCKN) relies on the 
successive uses of the Nystr\"om method~\citep{williams2001using}
to approximate the feature map at each layer, which makes our
approach scalable. GCKNs can then be interpreted as a new type of graph neural network whose
filters may be learned without supervision, by following kernel approximation principles. Such unsupervised graph
representation is known to be particularly effective when small
amounts of labeled data are available. Similar to CKNs, our model can also
be trained end-to-end, as a GNN, leading to task-adaptive representations, with a
computational complexity similar to that of a GNN when the path lengths are
small enough.

\paragraph{Notation.}
A graph $G$ is defined as a triplet $(\Vcal,\Ecal,a)$, where $\Vcal$
is the set of vertices, $\Ecal$ is the set of edges, and $a:\Vcal\to
\Sigma$ is a function that assigns attributes, either discrete or
continous, from a set $\Sigma$ to nodes in the graph.  A path is a
sequence of distinct vertices linked by edges and we denote by
$\Pcal(G)$ and $\Pcal_k(G)$ the set of paths and paths of length $k$
in $G$, respectively. In particular, $\Pcal_0(G)$ is reduced to
$\Vcal$. We also denote by $\Pcal_k(G, u)\subset\Pcal_k(G)$ the set of
paths of length~$k$ starting from $u$ in $\Vcal$.  For any path $p$ in
$\Pcal(G)$, we denote by $a(p)$ in $\Sigma^{|p|+1}$ the concatenation
of node attributes in this path. We replace $\Pcal$ with $\Wcal$ to
denote the corresponding sets of walks by allowing repeated nodes.

\section{Related Work on Graph Kernels}\label{sec:graphs}
Graph kernels were originally introduced by~\citet{gartner2003graph}
and~\citet{kashima2003marginalized}, and have been the subject of
intense research during the last twenty years~\citep[see the reviews
of][]{vishwanathan2010graph,kriege2019survey}.

In this paper, we consider graph kernels that represent a graph as a
feature vector counting the number of occurrences of some local
connected sub-structure.  Enumerating common local sub-structures
between two graphs is unfortunately often intractable; for instance,
enumerating common subgraphs or common paths is known to be
NP-hard~\cite{gartner2003graph}.  For this reason, the literature on
graph kernels has focused on alternative structures allowing for
polynomial-time algorithms, \eg, walks.

More specifically, we consider graph kernels that perform pairwise comparisons 
between local sub-structures centered at every node.
Given two graphs $G=(\Vcal, \Ecal, a)$ and $G'=(\Vcal',\Ecal',
a')$, we consider the kernel
\begin{equation}\label{eq:graph_kernel}
	K(G, G')=\sum_{u\in \Vcal} \sum_{u'\in \Vcal'} \kappa_{\text{base}}(l_G(u),l_{G'}(u')),
\end{equation}
where the base kernel $\kappa_{\text{base}}$ compares a set of local patterns centered at nodes~$u$ and $u'$, denoted by $l_{G}(u)$ and $l_{G'}(u')$, respectively. For simplicity, we will omit the notation $l_G(u)$ in the rest of the paper, and the base kernel will be simply written $\kappa_{\text{base}}(u,u')$ with an abuse of notation.
As noted by~\citet{lei2017deriving,kriege2019survey}, this class of kernels covers most of the examples mentioned in the introduction.

\paragraph{Walks and path kernels.}
Since computing all path co-occurences between graphs is NP-hard, it is
possible instead to consider paths of length~$k$, which can be reasonably 
enumerated if $k$ is small enough, or the graphs are sparse. Then,
we may define the kernel $K_{\text{path}}^{(k)}$ as~(\ref{eq:graph_kernel}) with
\begin{equation}
  \label{eq:path_base}
   \kappa_{\text{base}}(u,u') = \sum_{p\in\Pcal_k(G,u)} \sum_{p'\in\Pcal_k(G',u')} \delta(a(p), a'(p')),
\end{equation}
where
$a(p)$ represents the attributes for path~$p$ in $G$, and $\delta$ is the Dirac kernel such that $\delta(a(p), a'(p'))=1$ if
$a(p)=a'(p')$ and $0$ otherwise.

It is also possible to define a variant that enumerates all paths up to length~$k$, by simply adding the kernels~$K_\text{path}^{(i)}$:
 \begin{equation}
 	K_{\text{path}}(G,G')=\sum_{i=0}^k K_{\text{path}}^{(i)}(G, G').
 \end{equation}
Similarly, one may also consider using walks by simply replacing the notation $\Pcal$ by~$\Wcal$ in the
previous definitions.

\paragraph{Weisfeiler-Lehman subtree kernels.} A subtree is a subgraph with a tree structure. It can be extended to
subtree patterns~\citep{shervashidze2011weisfeiler,bach2008graph} by
allowing nodes to be repeated, just as the notion of walks extends
that of paths. All previous subtree kernels compare subtree patterns instead of subtrees.
Among them, the Weisfeiler-Lehman (WL) subtree kernel is one of the most widely used graph kernels to capture such patterns. It is essentially based on a mechanism to augment node attributes by iteratively aggregating and hashing the attributes of each node's neighborhoods. After $i$ iterations, we denote by $a_i$ the new node attributes for graph $G=(\Vcal,\Ecal,a)$, which is defined in Algorithm 1 of \citet{shervashidze2011weisfeiler} and then the WL subtree kernel after $k$ iterations is defined, for two graphs $G=(\Vcal,\Ecal,a)$ and $G'=(\Vcal',\Ecal',a')$, as
\begin{equation}
	K_{WL}(G, G')=\sum_{i=0}^k K_{\text{subtree}}^{(i)}(G, G'), \end{equation}
where 
\begin{equation}
  \label{eq:subtreekernel}
	K_{\text{subtree}}^{(i)} (G, G')=\sum_{u\in \Vcal} \sum_{u'\in \Vcal'} \kappa_{\text{subtree}}^{(i)}(u, u'),
\end{equation}
with $\kappa_{\text{subtree}}^{(i)}(u, u')=\delta(a_i(u), a_i'(u'))$ and 
the attributes $a_i(u)$ capture subtree patterns of depth~$i$ rooted at node~$u$.

\section{Graph Convolutional Kernel Networks}
\label{sec:gckn}
In this section, we introduce our model, which builds upon the concept of
graph-structured feature maps, following the terminology of convolutional
neural networks.  
\begin{definition}[Graph feature map]
   Given a graph $G=(\Vcal,\Ecal,a)$ and a RKHS $\Hcal$, a graph feature map is a mapping $\varphi: \Vcal \to \Hcal$, which associates to every node a point in $\Hcal$ representing information about local graph substructures.
\end{definition}
We note that the definition matches that of convolutional kernel
networks~\citep{mairal2016end} when the graph is a
two-dimensional grid. Generally, the map $\varphi$ depends
on the graph $G$, and can be seen as a collection of $|\Vcal|$
elements of~$\Hcal$ describing its nodes. The kernel associated to
the feature maps $\varphi, \varphi'$ for two graphs $G, G'$, is
defined as
\begin{equation}
   K(G,G') \!=\! \sum_{u \in \Vcal} \sum_{u' \in \Vcal'} \langle \varphi(u), \varphi'(u') \rangle_{\Hcal} \!=\!  \langle \Phi(G), \Phi(G') \rangle_{\Hcal} , \label{eq:kernel}
\end{equation}
with
\begin{equation}
   \Phi(G) =  \sum_{u \in \Vcal} \varphi(u)~~~\text{and}~~~\Phi(G') =  \sum_{u \in \Vcal'} \varphi'(u). \label{eq:featuremap}
 \end{equation}
 
The RKHS of~$K$ can be characterized by using
Theorem~\ref{thm:rkhs} in Appendix~\ref{sec:useful_res}. It is the space of
functions $f_z: G \mapsto \langle z, \Phi(G) \rangle_{\Hcal}$ for all
$z$ in~$\Hcal$ endowed with a particular norm.

Note that even though graph feature maps~$\varphi, \varphi'$ are graph-dependent, learning
with $K$ is possible as long as they all map nodes to the same RKHS
$\Hcal$---as $\Phi$ will then also map all graphs to the same space $\Hcal$.
We now detail the full construction of
the kernel, starting with a single layer.

\subsection{Single-Layer Construction of the Feature Map}
We propose a single-layer model corresponding to a continuous
relaxation of the path kernel. We assume that the input attributes
$a(u)$ live in~$\Real^{q_0}$, such that a graph $G=(\Vcal,\Ecal,a)$ admits a
graph feature map $\varphi_0: \Vcal \to \Hcal_0$ with
$\Hcal_0=\Real^{q_0}$ and $\varphi_0(u)=a(u)$. Note that this
assumption also allows us to handle discrete labels by using a one-hot
encoding strategy---that is \eg, four labels $\{A,B,C,D\}$ are represented
by four-dimensional vectors $(1,0,0,0),(0,1,0,0),(0,0,1,0),(0,0,0,1)$,
respectively.

\paragraph{Continuous relaxation of the path kernel.}
We rely on paths of length~$k$, and introduce the kernel~$K_1$ for graphs $G,G'$ with feature maps $\varphi_0,\varphi_0'$ of the
form~\eqref{eq:graph_kernel} with
\begin{equation}
  \label{eq:kbase}
   \kappa_{\text{base}}(u,u') =   \sum_{p\in\Pcal_k(G,u)}
   \sum_{p'\in\Pcal_k(G',u')} \kappa_1(\varphi_0(p),\varphi_0'(p')),   
\end{equation}
   where $\varphi_0(p) = [\varphi_0(p_i)]_{i=0}^k$ denotes the
concatenation of $k+1$ attributes along path~$p$, which is an element
of~$\Hcal_0^{k+1}$, $p_i$ is the $i$-th node on path~$p$ starting from
index $0$, and $\kappa_1$ is a Gaussian kernel comparing such
attributes:
\begin{equation}
   \kappa_1(\varphi_0(p),\varphi_0'(p')) = e^{-\frac{\alpha_1}{2} \sum_{i=0}^{k} \|\varphi_0(p_i) - \varphi_0'(p'_i)\|_{\Hcal_0}^2}.\label{eq:gaussian}
\end{equation}
This is an extension of the path kernel, obtained by replacing the
hard matching function $\delta$ in~\eqref{eq:path_base} by~$\kappa_1$,
   as done for instance by~\citet{togninalli2019wasserstein} for the WL kernel.
This replacement not only allows us to use continuous attributes, but
also has important consequences in the discrete case since it allows
to perform inexact matching between paths. For instance, when the
graph is a chain with discrete attributes---in other words, a
string---then, paths are simply $k$-mers, and the path kernel (with
matching function $\delta$) becomes the spectrum kernel for
sequences~\citep{leslie2001spectrum}. By using~$\kappa_1$ instead, we
obtain the single-layer CKN kernel of~\citet{chen2019bio}, which
performs inexact matching, as the mismatch kernel
does~\citep{leslie2004mismatch}, and leads to better performances in
many tasks involving biological sequences.

\begin{figure*}[ht]
  \centering
  \includegraphics[width=0.9\linewidth]{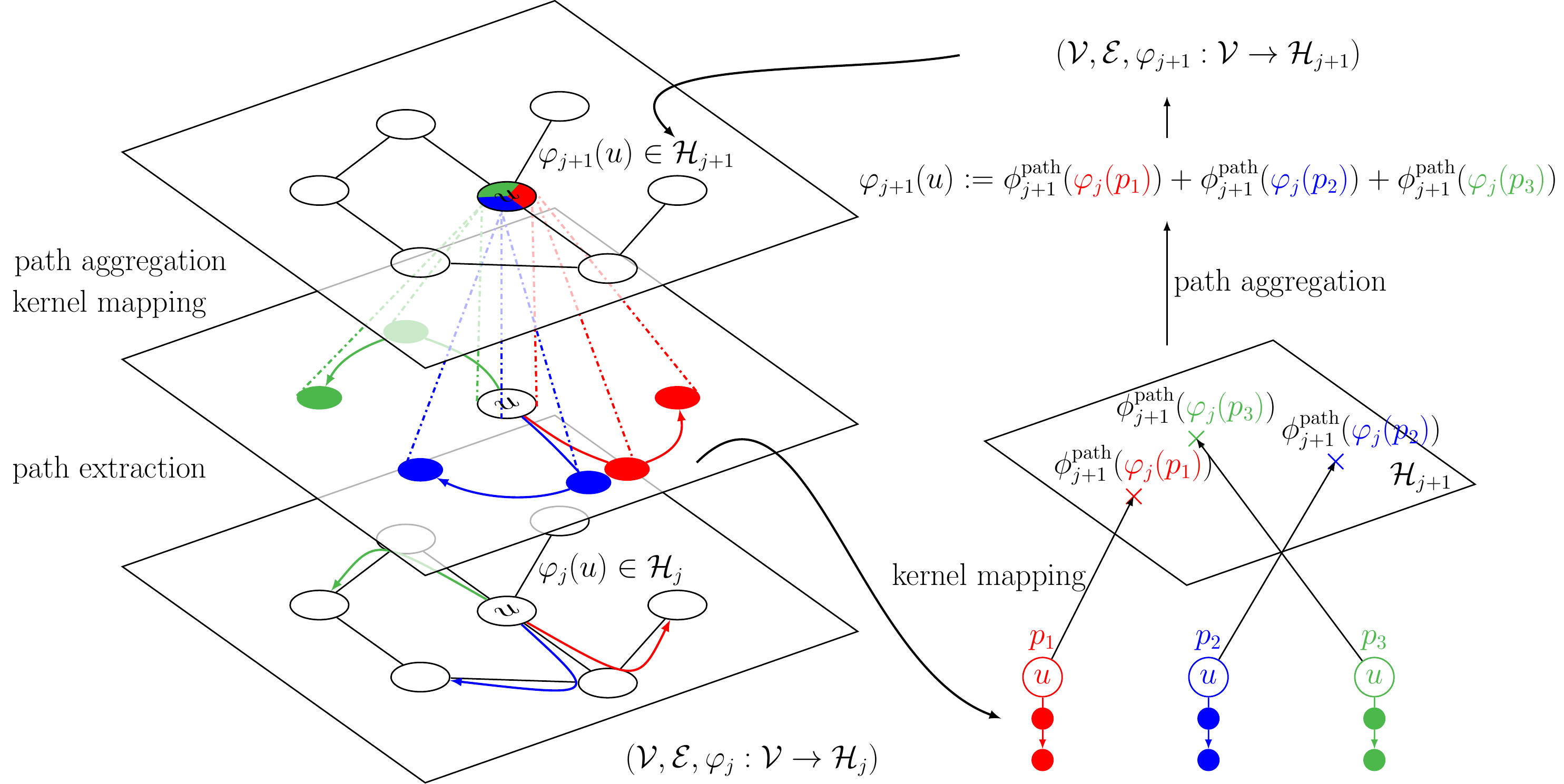}
   \caption{Construction of the graph feature map $\varphi_{j+1}$ from $\varphi_j$ given a graph $(\Vcal,\Ecal)$. The first step extracts paths of length~$k$ (here colored by red, blue and green) from node $u$, then (on the right panel) maps them to a RKHS~$\Hcal_{j+1}$ via the Gaussian kernel mapping. The new map~$\varphi_{j+1}$ at $u$ is obtained by local path aggregation (pooling) of their representations in $\Hcal_{j+1}$. The representations for other nodes can be obtained in the same way. In practice, such a model is implemented by using finite-dimensional embeddings approximating the feature maps, see Section~\ref{subsec:impl}.}
  \label{fig:gckn}
\end{figure*}

\paragraph{From graph feature map~$\varphi_0$ to graph feature
  map~$\varphi_1$.}
The kernel $\kappa_1$ acts on pairs of paths in potentially different
graphs, but only through their mappings to the same
space~$\Hcal_0^{k+1}$.  Since $\kappa_1$ is positive definite, we
denote by $\Hcal_1$ its RKHS and consider its 
mapping~$\phi_1^{\textrm{path}}:\Hcal_0^{k+1}\to\Hcal_1$ such that
$$\kappa_1(\varphi_0(p), \varphi_0'(p')) = \langle \phi_1^{\textrm{path}}\left(\varphi_0(p)\right),
\phi_1^{\textrm{path}}\left(\varphi_0'(p')\right)\rangle_{\Hcal_1}.$$
For any graph $G$, we can now define a graph feature map $\varphi_1:
\Vcal \to \Hcal_1$, operating on nodes $u$ in~$\Vcal$, as
\begin{equation}
  \label{eq:varphi1}
  \varphi_1(u) =  \sum_{p\in\Pcal_k(G,u)} \phi_1^{\textrm{path}}\left(\varphi_0(p)\right).
\end{equation}
Then, the continuous relaxation of the path kernel, denoted
by~$K_1(G,G')$, can also be written as~(\ref{eq:kernel}) with
$\varphi=\varphi_1$, and its underlying kernel representation~$\Phi_1$
is given by~(\ref{eq:featuremap}).  The construction of $\varphi_1$
from~$\varphi_0$ is illustrated in Figure~\ref{fig:gckn}.

The graph feature map $\varphi_0$ maps a node (resp a path) to
$\Hcal_0$ (resp $\Hcal_0^{k+1}$) which is typically a Euclidean space
describing its attributes. By contrast, $\phi_1^{\textrm{path}}$ is
the kernel mapping of the Gaussian kernel $\kappa_1$, and maps each path
$p$ to a Gaussian function centered at $\varphi_0(p)$---remember indeed
that for kernel function $K: \Xcal \times \Xcal \to \Real$ with RKHS~$\Hcal$,
the kernel mapping is of a data point $x$ is the function $K(x,.): \Xcal \to \Real$.
Finally,
$\varphi_1$ maps each node $u$ to a mixture of Gaussians, each Gaussian function
corresponding to a path starting at~$u$.

\subsection{Concrete Implementation and GCKNs}\label{subsec:impl}
We now discuss algorithmic aspects, leading to the graph convolutional kernel network (GCKN) model, which consists in
building a finite-dimensional embedding $\Psi(G)$  that
may be used in various learning tasks without scalability issues.
We start here with the single-layer case.

\paragraph{The Nystr\"om method and the single-layer model.}
 A naive computation of
   the path kernel~$K_1$ requires comparing all pairs of paths in each pair of graphs.
To gain scalability, a key component of the CKN model is the Nystr\"om method~\citep[][]{williams2001using}, which computes
finite-dimensional approximate kernel embeddings.  
   We discuss here the use of such a technique to define finite-dimensional maps $\psi_1: \Vcal \to \Real^{q_1}$ and $\psi_1': \Vcal' \to \Real^{q_1}$ for graphs~$G,G'$ such that for all pairs of nodes $u,u'$ in $\Vcal$, $\Vcal'$, respectively,
\begin{displaymath}
   \langle \varphi_1(u), \varphi_1'(u') \rangle_{\Hcal_1} \approx \langle \psi_1(u) , \psi_1'(u') \rangle_{\Real^{q_1}}.
\end{displaymath}
The consequence of such an approximation is that it provides a finite-dimensional approximation $\Psi_1$ of $\Phi_1$:
\begin{displaymath}
   \begin{split}
      K_1(G,G') & \approx \langle \Psi_1(G), \Psi_1(G') \rangle_{\Real^{q_1}} \\
      \text{with}~ \Psi_1(G) & = \sum_{u \in \Vcal} \psi_1(u).
   \end{split}
\end{displaymath}
Then, a supervised learning problem with kernel~$K_1$ on a dataset
$(G_i,y_i)_{i=1,\ldots,n}$, where $y_i$ are labels in $\Real$, can be solved
by minimizing the regularized empirical risk
\begin{equation}\label{eq:classifier}
  \min_{w\in\R^{q_1}} \sum_{i=1}^n L( y_i, \langle \Psi_1(G_i), w \rangle)+\lambda \| w\|^2,
\end{equation}
where $L$ is a convex loss function.
Next, we show that using the Nystr\"om method to approximate the kernel~$\kappa_1$
yields a new type of GNN, represented by $\Psi_1(G)$, whose filters can be
obtained without supervision, or, as discussed later, with back-propagation in
a task-adaptive manner.

Specifically, the Nystr\"om method projects points from a given RKHS
onto a finite-dimensional subspace and performs all subsequent
operations within that subspace.  In the context of $\kappa_1$, whose
RKHS is~$\Hcal_1$ with mapping function $\phi_1^\text{path}$, we
consider a collection $Z=\{z_1,\dots,z_{q_1}\}$ of~$q_1$ prototype
paths represented by attributes in $\Hcal_0^{k+1}$, and we define the
subspace
$\Ecal_1=\text{Span}(\phi_1^\text{path}(z_1),\dots,\phi_{1}^\text{path}(z_{q_1}))$.
Given a new path with attributes $z$, it is then possible to show
\citep[see][]{chen2019bio} that the projection of path attributes~$z$
onto~$\Ecal_1$ leads to the $q_1$-dimensional mapping
 \begin{equation*}
    \psi_1^\text{path}(z)= [\kappa_{1}(z_i,z_j)]_{ij}^{{-\frac{1}{2}}} [ \kappa_1(z_1,z), \ldots,  \kappa_1(z_{q_1},z)]^\top,
 \end{equation*}
where $[\kappa_{1}(z_i,z_j)]_{ij}$ is a $q_1\times q_1$ Gram matrix. 
Then, the approximate graph feature map~$\psi_1$ is obtained by pooling
\begin{equation*}
   \psi_1(u) =  \sum_{p\in\Pcal_k(G,u)} \psi_1^{\text{path}}( \psi_0(p))~~~~\text{for all}~~~~u \in \Vcal,
\end{equation*}
where $\psi_0\!=\!\varphi_0$ and $\psi_0(p) = [\psi_0(p_i)]_{i=0,\ldots,k}$ in $\Real^{q_0 (k+1)}$ represents the attributes of path~$p$, with an abuse of notation.

\paragraph{Interpretation as a GNN.}
When input attributes $\psi_0(u)$ have unit-norm, which is the case if we use one-hot encoding on discrete attributes, the Gaussian kernel $\kappa_1$ between two 
path attributes $z,z'$ in $\Real^{q_0 (k+1)}$ may be written
\begin{equation}
   \kappa_1(z,z') = e^{-\frac{\alpha_1}{2}\|z-z'\|^2} = e^{\alpha_1 (z^\top z' - k - 1)} = \sigma_1(z^\top z'),\label{eq:sigma0}
\end{equation}
which is a dot-product kernel with a non-linear function $\sigma_1$.
Then, calling $Z$ in $\Real^{q_0 (k+1)  \times q_1}$ the matrix of prototype path attributes, we have 
\begin{equation}
   \psi_1(u) =  \sum_{p\in\Pcal_k(G,u)} \sigma_1(Z^\top Z)^{-\frac{1}{2}}  \sigma_1(Z^\top \psi_0(p)),\label{eq:psi}
\end{equation}
where, with an abuse of notation, the non-linear function~$\sigma_1$ is applied pointwise.
Then, the map $\psi_1$ is build from $\psi_0$ with the following steps (i) feature aggregation along the paths,
(ii) encoding of the paths with a linear operation followed by point-wise non-linearity, (iii) multiplication by 
the $q_1 \times q_1$ matrix $\sigma_1(Z^\top Z)^{-\frac{1}{2}}$, and (iv) linear pooling.
The major difference with a classical GNN is that the ``filtering'' operation may be interpreted as an orthogonal 
projection onto a linear subspace, due to the matrix $\sigma_1(Z^\top Z)^{-\frac{1}{2}}$. Unlike the Dirac function, the exponential function $\sigma_1$ is differentiable. A useful consequence is the possibility of optimizing the filters $Z$ with back-propagation as detailed below.
Note that in practice we add a small regularization term to the diagonal for stability reason: $(\sigma_1(Z^\top Z)+\varepsilon I )^{-\frac{1}{2}}$ with $\varepsilon=0.01$.
\paragraph{Learning without supervision.}
Learning the ``filters'' $Z$ with Nystr\"om can be achieved by simply
running a K-means algorithm on path attributes extracted from training
data~\citep{zhang2008improved}.  This is the approach adopted for CKNs
by~\citet{mairal2016end,chen2019bio}, which proved to be very
effective as shown in the experimental section.

\paragraph{End-to-end learning with back-propagation.}
While the previous unsupervised learning strategy consists in finding
a good kernel approximation  that is independent of labels, it is 
also possible to learn the parameters~$Z$ end-to-end, by
minimizing~(\ref{eq:classifier}) jointly with respect to~$Z$ and~$w$.
The main observations from~\citet{chen2019bio} in the context of biological 
sequences is that such a supervised learning approach may yield  good models
with much fewer filters~$q_1$ than with the unsupervised learning strategy.
We refer the reader to~\citet{chen2019bio,chen2019rec} for how to perform
back-propagation with the inverse square root matrix  $\sigma_1(Z^\top Z)^{-\frac{1}{2}}$. 

\paragraph{Complexity.}
The complexity for computing the feature map~$\psi_1$ is dominated by the complexity
of finding all the paths of length $k$ from each node. This can be done by
simply using a depth first search algorithm, whose worst-case complexity for
each graph is $O(|\Vcal| d^k)$, where $d$ is the maximum degree of each node,
meaning that large~$k$ may be used only for sparse graphs. Then, each path is encoded
in $O(q_1 q_0 (k+1))$ operations;
When learning with back-propagation, each gradient step requires computing the
eigenvalue decomposition of $\sigma_1(Z^\top Z)^{-\frac{1}{2}}$  whose complexity is
$O(q_1^3)$, which is not a computational bottleneck when using mini-batches of order $O(q_1)$, where typical practical values for $q_1$ are reasonably small, \eg, less than $128$.

\subsection{Multilayer Extensions}\label{sec:multilayer}
The mechanism to build the feature map~$\varphi_1$ from~$\varphi_0$
can be iterated, as illustrated in Figure~\ref{fig:gckn} which shows
how to build a feature map~$\varphi_{j+1}$ from a previous
one~$\varphi_j$. As discussed by~\citet{mairal2016end} for CKNs, the
Nystr\"om method may then be extended to build a sequence of
finite-dimensional maps $\psi_0,\ldots,\psi_J$, and the final graph
representation is given by
\begin{equation}\label{eq:Psi}
	\Psi_J(G) = \sum_{u \in \Vcal} \psi_J(u).
\end{equation}

The computation of $\Psi_J(G)$ is illustrated in Algorithm~\ref{alg:forward}.
Here we discuss two possible uses for these additional layers, either
to account for more complex structures than paths, or to extend the
receptive field of the representation without resorting to the
enumeration of long paths.We will denote by $k_j$ the path length used at
layer~$j$.

\begin{algorithm}[bt]
	\caption{Forward pass for multilayer GCKN}
	\label{alg:forward}
	\begin{algorithmic}[1]
           \STATE {\bfseries Input:} graph $G=(\Vcal,\Ecal,\psi_0:\Vcal\to \R^{q_0})$, set of anchor points (filters) $Z_j\in\R^{(k+1) q_{j-1}\times q_j}$ for $j=1,\ldots,J$.
   		\FOR{$j=1,\ldots,J$}
   		\FOR{$u$ {\bfseries in} $\Vcal$}
                \STATE $\psi_j(u)=\sum_{p\in\Pcal_k(G,u)} \psi_j^{\text{path}}( \psi_{j-1}(p))$;
   		\ENDFOR
                \ENDFOR
   		\STATE Global pooling: $\Psi(G)=\sum_{u\in\Vcal}\psi_J(u)$;
	\end{algorithmic}
\end{algorithm}

\paragraph{A simple two-layer model to account for subtrees.}

\begin{figure}[ht]
  \centering
  \includegraphics[width=\linewidth]{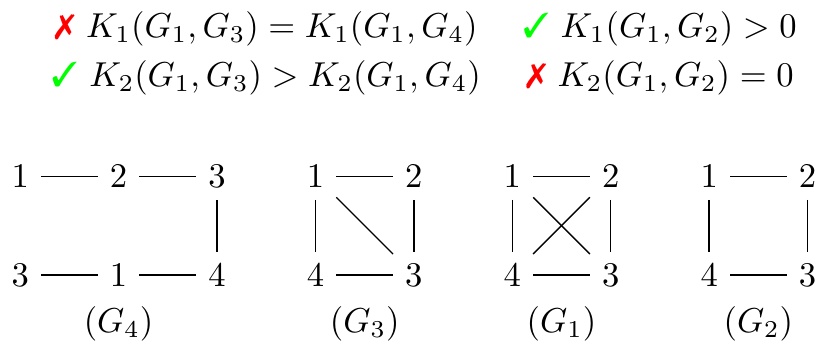}
  \caption{Example cases using $\kappa_1=\kappa_2=\delta$, with path
    lengths $k_1=1$ and $k_2=0$; The one-layer kernel $K_1$ counts the
    number of common edges while the two-layer $K_2$ counts the number
    of nodes with the same set of outgoing edges. The figure suggests
    using $K_1+K_2$ to gain expressiveness. }
  \label{fig:example}
\end{figure}

As emphasized in~\eqref{eq:featuremap}, GCKN relies on a
representation $\Phi(G)$ of graphs, which is a sum of node-level
representations provided by a graph feature map $\varphi$. If
$\varphi$ is a sum over paths starting at the represented node,
$\Phi(G)$ can simply be written as a sum over all paths in $G$,
consistently with our observation that~\eqref{eq:kernel} recovers the
path kernel when using a Dirac kernel to compare paths in
$\kappa_1$. The path kernel often leads to good performances,
but it is also blind to more complex
structures. Figure~\ref{fig:example} provides a simple example of this
phenomenon, using $k=1$: $G_1$ and $G_3$ differ by a single edge,
while $G_4$ has a different set of nodes and a rather different
structure. Yet $\Pcal_1(G_3) = \Pcal_1(G_4)$, making
$K_1(G_1, G_3) = K_1(G_1, G_4)$ for the path kernel.

Expressing more complex structures requires breaking the succession of
linearities introduced in~\eqref{eq:featuremap}
and~\eqref{eq:varphi1}---much like pointwise nonlinearities are used
in neural networks. Concretely, this effect can simply be obtained
by using a second layer with path length $k_2=0$---paths are then
identified to vertices---which produces
the feature map $\varphi_2(u) = \phi_2^{\text{path}}(\varphi_1(u))$,
where $\phi_2^{\text{path}} : \Hcal_1 \to \Hcal_2$ is a non-linear kernel mapping.
The resulting kernel is then
\begin{align}
  \label{eq:stratifiedkernel}
 K_2(G, G') &= \sum_{u \in \Vcal} \sum_{u' \in \Vcal'} \langle
            \varphi_2(u), \varphi'_2(u') \rangle_{\Hcal_2}\nonumber\\
          &= \sum_{u \in \Vcal}
\sum_{u' \in \Vcal'} \kappa_2(\varphi_1(u), \varphi'_1(u')).
\end{align}
When $\kappa_1$ and $\kappa_2$ are both Dirac kernels, $K_2$ counts
the number of nodes in $G$ and $G'$ with the exact same set of
outgoing paths $\Pcal(G,u)$, as illustrated in
Figure~\ref{fig:example}.

Theorem~\ref{thm:subtree_walk} further illustrates the effect of using
a nonlinear~$\phi_2^\text{path}$ on the feature map~$\varphi_1$,
by formally linking the walk and WL subtree kernel through our
framework. 
\begin{theorem}\label{thm:subtree_walk}
  Let $G=(\Vcal, \Ecal),\, G'=(\Vcal', \Ecal')$, $\Mcal$ be the set of
  exact matchings of subsets of the neighborhoods of two nodes, as
  defined in~\citet{shervashidze2011weisfeiler}, and $\varphi$ defined
  as in~\eqref{eq:varphi1} with $\kappa_1=\delta$ and replacing paths
  by walks. For any $u\in\Vcal$ and $u'\in\Vcal'$ such that
  $|\Mcal(u,u')|= 1$,
  \begin{equation}\label{eq:link_subtree_walk}
    \delta(\varphi_1(u),\varphi_1'(u')) = \kappa_{\text{subtree}}^{(k)}(u,u').
  \end{equation}
\end{theorem}
Recall that when using~\eqref{eq:kbase} with walks instead
of paths and a Dirac kernel for $\kappa_1$, the
kernel~\eqref{eq:kernel} with $\varphi=\varphi_1$ is the walk kernel.
The condition $|\Mcal(u,u')|= 1$ indicates that $u$ and $u'$ have the
same degrees and each of them has distinct neighbors. This can be always
ensured by including degree information and adding noise
to node attributes. For a large class of graphs, both the walk and WL
subtree kernels can therefore be written as~\eqref{eq:kernel} with the
same first layer $\varphi_1$ representing nodes by their walk
histogram. While walk kernels use a single layer, WL subtree kernels rely on a
second layer~$\varphi_2$ mapping nodes to the indicator function of
$\varphi_1(u)$.

Theorem~\ref{thm:subtree_walk} also shows that the kernel built
in~\eqref{eq:stratifiedkernel} is a path-based version of WL subtree
kernels, therefore more expressive as it captures subtrees rather than
subtree patterns. However, the Dirac kernel lacks flexibility, as it
only accounts for pairs of nodes with identical $\Pcal(G, u)$. For
example, in Figure~\ref{fig:example}, $K_2(G_1,G_2)=0$ even though
$G_1$ only differs from~$G_2$ by two edges, because these two edges
belong to the set $\Pcal(G, u)$ of all nodes in the graph. In order to
retain the stratification by node of~\eqref{eq:stratifiedkernel} while
allowing for a softer comparison between sets of outgoing paths, we
replace $\delta$ by the kernel
$\kappa_2(\varphi_1(u),\varphi_1'(u')) = e^{-\alpha_2\|\varphi_1(u) -
  \varphi'_1(u')\|_{\Hcal_1}^2}$.
Large values of $\alpha_2$ recover the
behavior of the Dirac, while smaller values gives non-zero values for
similar $\Pcal(G,u)$.

\paragraph{A multilayer model to account for longer paths.}
In the previous paragraph, we have seen that adding a second layer could bring
some benefits in terms of expressiveness, even when using path lengths $k_2=0$.
Yet, a major limitation of this model is the exponential complexity of path enumeration,
which is required to compute the feature map~$\varphi_1$, preventing us to 
use large values of $k$ as soon as the graph is dense.
Representing large receptive fields while relying on path enumerations with small $k$, \eg, $k \leq 3$,
is nevertheless possible with a multilayer model.
To account for a receptive field of size $k$, the previous model requires a
path enumeration with complexity $O(|\Vcal|d^k)$, whereas the complexity of a multilayer model
is linear in $k$.

\subsection{Practical Variants}
\paragraph{Summing the kernels for different~$k$ and different scales.}
As noted in Section~\ref{sec:graphs}, summing the kernels corresponding to
different values of~$k$ provides a richer representation.
We also adopt such a strategy, which corresponds to concatenating the 
feature vectors $\Psi(G)$ obtained for various path lengths~$k$.
When considering a multilayer model, it is also possible to concatenate 
the feature representations obtained at every layer~$j$, allowing to
obtain a multi-scale feature representation of the graph and gain expressiveness.

\paragraph{Use of homogeneous dot-product kernel.}
Instead of the Gaussian kernel~(\ref{eq:gaussian}), it is possible to use a homogeneous dot-product kernel, as suggested by~\citet{mairal2016end} for CKNs:
\begin{displaymath}
   \kappa_1(z,z') = \|z\| \|z'\| \sigma_1\left ( \frac{\langle z, z'\rangle}{\|z\| \|z'\|}\right),
\end{displaymath}
where $\sigma_1$ is defined in~(\ref{eq:sigma0}). Note that when $z,z'$ have unit-norm, we recover the Gaussian kernel~(\ref{eq:gaussian}). In our paper, we use such a kernel for upper layers, or for continuous input attributes when they do not have unit norm. For multilayer models, this homogenization is useful for preventing vanishing or exponentially growing representations. Note that ReLU is also a homogeneous non-linear mapping.

\paragraph{Other types of pooling operations.}
Another variant consists in replacing the sum pooling operation in~\eqref{eq:psi} and \eqref{eq:Psi}
by a mean or a max pooling. While using max pooling as a heuristic seems to be effective on some datasets,
it is hard to justify from a RKHS point of view since max operations typically
do not yield positive definite kernels. Yet, such a heuristic is widely adopted
in the kernel literature, \eg, for string alignment kernels~\citep{saigo2004protein}.
In order to solve such a discrepancy between theory and practice, \citet{chen2019rec} propose to use the generalized max pooling operator
of~\citet{murray2014generalized}, which is compatible with the RKHS point of
view. Applying the same ideas to GCKNs is straightforward.

\paragraph{Using walk kernel instead of path kernel.}
One can use a relaxed walk kernel instead of the path kernel in~\eqref{eq:kbase}, at the cost of losing some expressiveness but gaining some time complexity. Indeed, there exists a very efficient recursive way to enumerate walks and thus to compute the resulting approximate feature map in~\eqref{eq:psi} for the walk kernel. Specifically, if we denote the $k$-walk kernel by $\kappa_{\text{walk}}^{(k)}$, then its value between two nodes can be decomposed as the product of the $0$-walk kernel between the nodes and the sum of the $(k-1)$-walk kernel between their neighbors
\begin{equation*}
	\kappa_{\text{walk}}^{(k)}(u,u')=\kappa_{\text{walk}}^{(0)}(u,u')\sum_{v\in\Ncal(u)} \sum_{v'\in\Ncal(u')} \kappa_{\text{walk}}^{(k-1)}(v, v'),
\end{equation*} 
where $\kappa_{\text{walk}}^{(0)}(u,u')=\kappa_1(\varphi_0(u),\varphi_0'(u') )$. After applying the Nystr\"om method, the approximate feature map of the walk kernel is written, similar to~\eqref{eq:psi}, as
\begin{equation*}
	\psi_1(u)= \sigma_1(Z^\top Z)^{-\frac{1}{2}}\underbrace{\sum_{p\in\Wcal_k(G,u)}  \sigma_1(Z^\top \psi_0(p))}_{c_k(u):=}.
\end{equation*}
Based on the above observation and following similar induction arguments as~\citet{chen2019rec}, it is not hard to show that $(c_j(u))_{j=1,\dots,k}$ obeys the following recursion
\begin{equation*}
	c_j(u)=b_j(u)\odot \sum_{v\in\Ncal(u)} c_{j-1}(v),~ 1\leq j\leq k,
\end{equation*}
where $\odot$ denotes the element-wise product and $b_j(u)$ is a vector in $\Real^{q_1}$ whose entry $i$ in $\{1,\dots,q_1\}$ is $\kappa_1(u,z_{i}^{(k+1-j)})$ and $z_{i}^{(k+1-j)}$ denotes the $k+1-j$-th column vector of $z_i$ in $\Real^{q_0}$. More details can be found in Appendix~\ref{sec:variants}.

\section{Model Interpretation}
\label{sec:interpretation}

\citet{ying2019gnnexplainer} introduced an approach to interpret
trained GNN models, by finding a subgraph of an input graph $G$
maximizing the mutual information with its predicted label (note that
this approach depends on a specific input graph).
We show here how to adapt similar ideas to our framework.

\paragraph{Interpreting GCKN-path and GCKN-subtree.}
We call GCKN-path our model~$\Psi_1$ with a single layer, and GCKN-subtree our model~$\Psi_2$
with two layers but with $k_2=0$, which is the first model presented in
Section~\ref{sec:multilayer} that accounts for subtree structures.
As these models are built
upon path enumeration, we extend the method of~\citet{ying2019gnnexplainer} by identifying a
small subset of paths in an input graph $G$ preserving the
prediction. 
We then reconstruct a subgraph by merging the selected
paths. For simplicity, let us consider a one-layer model. As
$\Psi_1(G)$ only depends on $G$ through its set of paths $\Pcal_k(G)$,
we note $\Psi_1(\Pcal)$ with an abuse of notation for any subset of
$\Pcal$ of paths in $G$, to emphasize the dependency in this set of
paths. For a trained model $(\Psi_1,w)$ and a graph $G$, our objective
is to solve
\begin{equation}
	\min_{\Pcal'\subseteq \Pcal_k(G)} L(\hat{y},\langle \Psi_1(\Pcal'), w\rangle ) + \mu |\Pcal'|,
\end{equation}
where $\hat y$ is the predicted label of $G$ and $\mu$ a
regularization parameter controlling the number of paths to
select. This problem is combinatorial and can be computationally
intractable when $\Pcal(G)$ is
large. Following~\citet{ying2019gnnexplainer}, we relax it by using a
mask $M$ with values in $[0;1]$ over the set of paths, and replace the number of paths $|\Pcal'|$
by the $\ell_1$-norm of~$M$, which is known to have a sparsity-inducing effect~\citep{tibshirani1996regression}.
The problem then becomes
\begin{equation}
   \min_{M\in [0;1]^{|\Pcal_k(G)|}} L(\hat y,\langle \Psi_1(\Pcal_k(G)\odot M ), w\rangle ) + \mu \|M\|_{1}, \label{eq:pruning}
\end{equation}
where $\Pcal_k(G)\odot M$ denotes the use of $M(p)a(p)$ instead of
$a(p)$ in the computation of $\Psi_1$ for all $p$ in $\Pcal_k(G)$.
Even though the problem is non-convex due to the non-linear mapping $\Psi_1$, it may 
still be solved approximately by using projected gradient-based optimization techniques.

\paragraph{Interpreting multilayer models.}
By noting that $\Psi_j(G)$ only depends on the union of the set of paths $\Pcal_{k_l}(G)$, for all layers $l \leq j$, we introduce a collection of masks $M_l$ at each layer, and then optimize the same objective as~(\ref{eq:pruning}) over all masks~$(M_l)_{l=1,\ldots,j}$, with the regularization $\sum_{l=1}^j \|M_l\|_1$.
 
\section{Experiments}
\begin{table*}[ht]
\centering
\caption{Classification accuracies on graphs with discrete node attributes. The accuracies of other models are taken from \citet{du2019graph} except LDP, which we evaluate on our splits and for which we tune bin size, the regularization parameter in the SVM and Gaussian kernel bandwidth. Note that RetGK uses a different protocol, performing 10-fold cross-validation 10 times and reporting the average accuracy.}\label{tab:discrete}
	\resizebox{.9\textwidth}{!}{
	\begin{tabular}{lccccccc}
		\toprule
		Dataset & MUTAG & PROTEINS & PTC  & NCI1 & IMDB-B & IMDB-M & COLLAB \\ \midrule
		size    & 188   & 1113     & 344  & 4110 & 1000   & 1500   & 5000   \\
		classes & 2     & 2        & 2    & 2    & 2      & 3      & 3      \\
		avg $\sharp$nodes & 18 & 39 & 26   & 30   & 20    & 13     & 74     \\
		avg $\sharp$edges & 20 & 73 & 51   & 32   & 97    & 66     & 2458   \\ \midrule
		LDP & $88.9\pm 9.6$ & $73.3\pm 5.7$ & $63.8\pm 6.6$ & $72.0\pm 2.0$ & $68.5\pm 4.0$ & $42.9\pm 3.7$ & $76.1\pm 1.4$ \\ \midrule
		WL subtree & $90.4\pm 5.7$ &  $75.0\pm 3.1$ &  $59.9\pm 4.3$ & $\mathbf{86.0\pm 1.8}$ & $73.8\pm 3.9$ & $50.9\pm 3.8$ &  $78.9\pm 1.9$ \\
		AWL      & $87.9\pm 9.8$ & - & - & - & $74.5\pm 5.9$ & $51.5\pm 3.6$ & $73.9\pm 1.9$ \\ 
		RetGK      & $90.3\pm 1.1$ & $75.8\pm 0.6$ & $62.5\pm 1.6$ & $84.5\pm 0.2$ & $71.9\pm 1.0$ & $47.7\pm 0.3$ & $81.0\pm 0.3$ \\ 
		GNTK       &  $90.0 \pm 8.5$ & $75.6 \pm 4.2$ & $67.9 \pm 6.9$ & $84.2 \pm 1.5$ & $76.9 \pm 3.6$ & $52.8 \pm 4.6$ & $\mathbf{83.6 \pm 1.0}$ \\ \midrule
		GCN        & $85.6\pm 5.8$ & $76.0\pm 3.2$ & $64.2\pm 4.3$ & $80.2\pm 2.0$ &  $ 74.0\pm 3.4$ & $51.9\pm 3.8$ & $79.0\pm 1.8$ \\
		PatchySAN  & $92.6\pm 4.2$ & $ 75.9\pm 2.8$ & $60.0\pm 4.8$ & $78.6\pm 1.9$ & $71.0\pm 2.2$ & $45.2\pm 2.8$ & $72.6\pm 2.2$ \\
		GIN        & $89.4 \pm 5.6$ & $76.2 \pm 2.8$ & $64.6 \pm 7.0$ & $82.7 \pm 1.7$ & $75.1 \pm 5.1$ & $52.3 \pm 2.8$ & $80.2 \pm 1.9$ \\ \midrule
		GCKN-walk-unsup    & $92.8\pm 6.1$ & $75.7\pm 4.0
		$ & $65.9\pm 2.0$ & $80.1\pm 1.8$ & $75.9\pm 3.7$ & $53.4\pm 4.7$ & $81.7\pm 1.4$ \\
		GCKN-path-unsup      & $92.8\pm 6.1$ & $76.0\pm 3.4$ & $67.3\pm 5.0$ & $81.4\pm 1.6$ & $75.9\pm 3.7$ & $53.0\pm 3.1$ & $82.3\pm 1.1$ \\
		GCKN-subtree-unsup     & $95.0\pm 5.2$ & $\mathbf{76.4\pm 3.9}$ & $\mathbf{70.8\pm 4.6}$ & $83.9\pm 1.6$ & $\mathbf{77.8\pm 2.6}$ & $\mathbf{53.5\pm 4.1}$ & $83.2\pm 1.1$ \\
		GCKN-3layer-unsup & $\mathbf{97.2\pm 2.8}$ & $75.9\pm 3.2$ & $69.4\pm 3.5$ & $83.9\pm 1.2$ & $77.2\pm 3.8$ & $53.4\pm 3.6$ & $83.4\pm 1.5$ \\ \midrule
		GCKN-subtree-sup       & $91.6\pm 6.7$ & $76.2\pm 2.5$ & $68.4\pm 7.4$  & $82.0\pm 1.2$ & $76.5\pm 5.7$ & $53.3\pm 3.9$ & $82.9\pm 1.6$ \\
		\bottomrule
	\end{tabular}
	}

\end{table*}
We evaluate GCKN and compare its variants to state-of-the-art methods, including GNNs and graph kernels, on several real-world graph classification datasets, involving either discrete or continuous attributes.

\subsection{Implementation Details}
We follow the same protocols as~\citep{du2019graph,xu2018how}, and
report the average accuracy and standard deviation over a 10-fold
cross validation on each dataset. We use the same data splits
as~\citet{xu2018how}, using their code. Note that performing nested 10-fold
cross validation would have provided better estimates of test accuracy 
for all models, but it would have unfortunately required 10 times more computation,
which we could not afford for many of the baselines we considered.

\paragraph{Considered models.}
We consider two single-layer models called GCKN-walk and GCKN-path,
corresponding to the continuous relaxation of the walk and path
kernels respectively. We also consider the two-layer model
GCKN-subtree introduced in Section~\ref{sec:multilayer} with path
length $k_2=0$, which accounts for subtrees. Finally, we consider a
3-layer model GCKN-3layers with path length $k_2=2$ (which enumerates
paths with three vertices for the second layer), and $k_3\!=\!0$, which
introduces a non-linear mapping before global pooling, as in
GCKN-subtree.  We use the same parameters $\alpha_j$ and $q_j$ (number
of filters) across layers.  Our comparisons include state-of-the-art
graph kernels such as WL kernel~\citep{shervashidze2011weisfeiler},
AWL~\citep{ivanov2018anonymous}, RetGK~\citep{zhang2018retgk},
GNTK~\citep{du2019graph}, WWL~\citep{togninalli2019wasserstein} and recent GNNs including
GCN~\citep{kipf2016semi}, PatchySAN~\citep{niepert2016learning} and
GIN~\citep{xu2018how}. We also include a simple baseline method LDP~\citep{cai2018simple} based on node degree information and a Gaussian SVM.

\paragraph{Learning unsupervised models.}
Following~\citet{mairal2016end}, we learn
the anchor points~$Z_j$ for each layer by K-means over 300000
extracted paths from each training fold. The resulting graph
representations are then mean-centered, standardized, and used within
a linear SVM classifier~\eqref{eq:classifier} with squared hinge loss. 
In practice, we use the SVM implementation of the Cyanure toolbox~\cite{mairal2019cyanure}.\footnote{\url{http://julien.mairal.org/cyanure/}}
For
each 10-fold cross validation, we tune the bandwidth of the Gaussian
kernel (identical for all layers), pooling operation
(local~\eqref{eq:psi} or global~\eqref{eq:Psi}), path size $k_1$ at
the first layer, number of filters (identical for all layers) and
regularization parameter~$\lambda$ in~\eqref{eq:classifier}. More
details are provided in Appendix~\ref{sec:add_exp}, as well as a study of the
model robustness to hyperparameters.

\paragraph{Learning supervised models.}
Following~\citet{xu2018how}, we use an Adam
optimizer~\citep{kingma2014adam} with the initial learning rate equal
to 0.01 and halved every 50 epochs, and fix the batch size to 32. We
use the unsupervised model based described above for initialization.
We select the best model based on the same hyperparameters as for
unsupervised models, with the number of epochs as an additional
hyperparameter as used in \citet{xu2018how}. Note that we do not use
DropOut or batch normalization, which are typically used in GNNs such
as \citet{xu2018how}. Importantly, the number of filters needed for
supervised models is always much smaller (\eg, 32 vs 512) than that for
unsupervised models to achieve comparable performance.

\subsection{Results}
\paragraph{Graphs with categorical node labels}
We use the same benchmark datasets as in \citet{du2019graph},
including 4 biochemical datasets MUTAG, PROTEINS, PTC and NCI1 and 3
social network datasets IMDB-B, IMDB-MULTI and COLLAB. All the
biochemical datasets have categorical node labels while none of the
social network datasets has node features. We use degrees as node
labels for these datasets, following the protocols of previous
works~\citep{du2019graph,xu2018how,togninalli2019wasserstein}. Similarly,
we also transform all the categorical node labels to one-hot
representations. The results are reported in
Table~\ref{tab:discrete}. With a few exceptions, GCKN-walk has a small
edge on graph kernels and GNNs---both implicitly relying on walks
too---probably because of the soft structure comparison allowed by the
Gaussian kernel. GCKN-path often brings some further improvement,
which can be explained by its increasing the expressivity. Both
multilayer GCKNs bring a stronger increase, whereas supervising the
filter learning of GCKN-subtree does not help.  Yet, the number of
filters selected by GCKN-subtree-sup is smaller than
GCKN-subtree-unsup (see Appendix~\ref{sec:add_exp}), allowing for 
faster classification at test time.  GCKN-3layers-unsup performs in
the same ballpark as GCKN-subtree-unsup, but benefits from lower
complexity due to smaller path length~$k_1$.

\paragraph{Graphs with continuous node attributes}
We use 4 real-world graph classification datasets with continuous node
attributes: ENZYMES, PROTEINS\_full, BZR, COX2. All datasets and size
information about the graphs can be found in~\citet{KKMMN2016}. The
node attributes are preprocessed with standardization as
in~\citet{togninalli2019wasserstein}. To make a fair comparison, we follow the
same protocol as used in~\citet{togninalli2019wasserstein}. Specifically, we 
perform 10 different 10-fold cross validations, using the same hyperparameters 
that give the best average validation accuracy. The hyperparameter search grids remain
the same as for training graphs with categorical node labels. The
results are shown in Table~\ref{tab:continuous}. They are comparable
to the ones obtained with categorical attributes, except that in 2/4
datasets, the multilayer versions of GCKN underperform compared to
GCKN-path, but they achieve lower computational complexity.  Paths
were indeed presumably predictive enough for these datasets. Besides,
the supervised version of GCKN-subtree outperforms its unsupervised
counterpart in 2/4 datasets.

\begin{table}[ht]
   \caption{Classification accuracies on graphs with continuous
     attributes. The accuracies of other models except GNTK are taken
     from~\citet{togninalli2019wasserstein}. The accuracies of GNTK
     are obtained by running the code of~\citet{du2019graph} on a
     similar setting.}\label{tab:continuous}
\centering
\resizebox{\columnwidth}{!}{
	\begin{tabular}{lccccc}
		\toprule
		Dataset & ENZYMES   & PROTEINS & BZR  & COX2       \\ \midrule
		size        & 600   & 1113    & 405   & 467        \\
		classes     & 6     & 2       & 2     & 2          \\
		attr.\ dim. & 18    & 29      & 3     & 3          \\ 
		avg $\sharp$nodes & 32.6 & 39.0 & 35.8 & 41.2      \\
		avg $\sharp$edges & 62.1 & 72.8 & 38.3 & 43.5      \\ \midrule
		RBF-WL      & $68.4\pm 1.5$  & $75.4\pm 0.3$   & $81.0\pm 1.7$ & $75.5\pm 1.5$    \\
		HGK-WL      & $63.0\pm 0.7$  & $75.9\pm 0.2$  & $78.6\pm 0.6$ & $78.1\pm 0.5$    \\
		HGK-SP      & $66.4\pm 0.4$  & $75.8\pm 0.2$   & $76.4\pm 0.7$ & $72.6\pm 1.2$    \\
		WWL         & $73.3\pm 0.9$  & $\mathbf{77.9\pm 0.8}$   & $84.4\pm 2.0$ & $78.3\pm 0.5$    \\ 
		GNTK      & $69.6\pm 0.9$ & $75.7\pm 0.2$ & $85.5\pm 0.8$ & $79.6\pm 0.4$ \\ \midrule
		GCKN-walk-unsup   & $73.5\pm 0.5$    & $76.5\pm 0.3$  & $85.3\pm 0.5$  & $80.6\pm 1.2$ \\
		GCKN-path-unsup   &   $\mathbf{75.7\pm 1.1}$ & $76.3\pm 0.5$ & $85.9\pm 0.5$ & $81.2\pm 0.8$ \\
		GCKN-subtree-unsup & $74.8\pm 0.7$ & $77.5\pm 0.3$ & $85.8\pm 0.9$ & $81.8\pm 0.8$ \\ 
		GCKN-3layer-unsup & $74.6\pm 0.8$ & $77.5\pm 0.4$ & $84.7\pm 1.0$ & $\mathbf{82.0\pm 0.6}$ \\ \midrule
		GCKN-subtree-sup & $72.8\pm 1.0$ & $77.6\pm 0.4$ & $\mathbf{86.4\pm 0.5}$ & $81.7\pm 0.7$ \\
		\bottomrule
	\end{tabular}
}
\end{table}

\subsection{Model Interpretation}
\begin{figure}[h!]
	\centering
	\includegraphics[width=0.99\columnwidth]{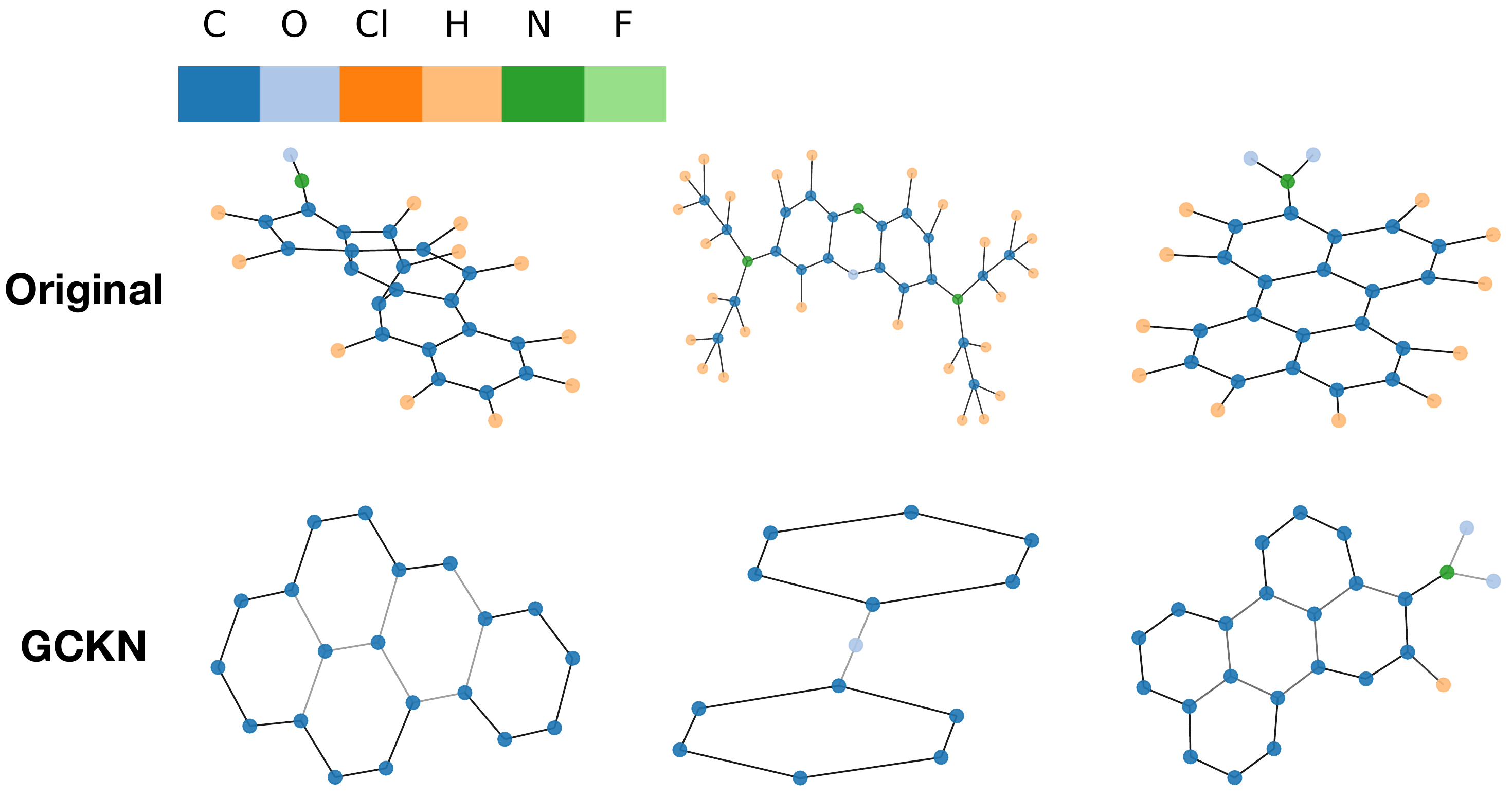} \\ \vspace{2.5mm}
	\includegraphics[width=0.99\columnwidth]{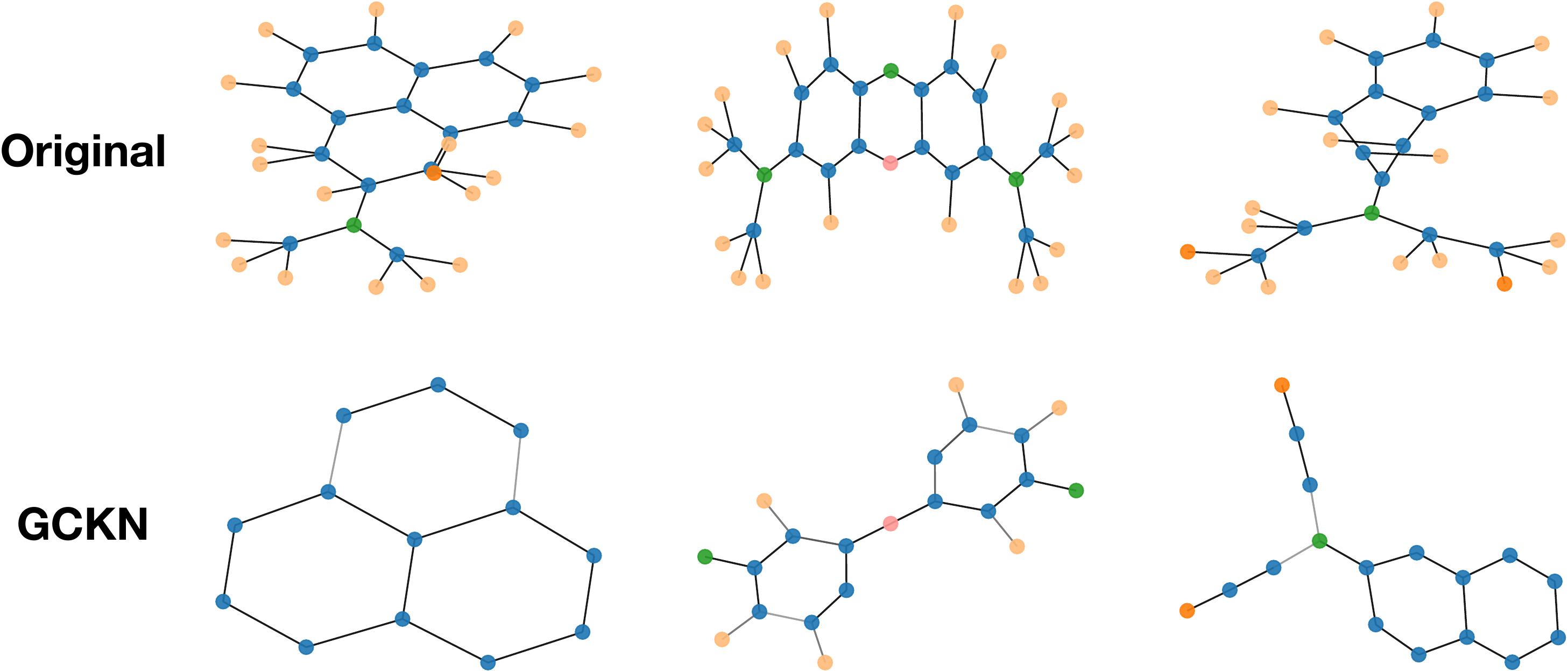}
	\caption{Motifs extracted by GCKN on the Mutagenicity dataset.}\label{fig:motif}
\end{figure}

We train a supervised GCKN-subtree model on the Mutagenicity dataset~\citep{KKMMN2016}, and use
our method described in Section~\ref{sec:interpretation} to identify
important subgraphs.  Figure~\ref{fig:motif} shows examples of detected subgraphs.
Our method is able to identify chemical groups known for
their mutagenicity such as Polycyclic aromatic hydrocarbon (top row left),
Diphenyl ether (top row middle) or $\text{NO}_2$ (top row right), thus admitting 
simple model interpretation. 
We also find some groups whose mutagenicity is not known, such as polyphenylene 
sulfide (bottom row middle) and 2-chloroethyl- (bottom row right).
More details and additional results are provided in Appendix~\ref{sec:add_exp}.

\section*{Acknowledgements}
This work has been supported by the grants from ANR (FAST-BIG project ANR-17 CE23-0011-01), by the ERC grant number 714381 (SOLARIS), and by ANR 3IA MIAI@Grenoble Alpes, (ANR-19-P3IA-0003).

\bibliography{mybib}
\bibliographystyle{icml2020}

\newpage
\appendix
\onecolumn

\vspace*{0.3cm}
\begin{center}
   {\huge Appendix}
\end{center}
\vspace*{0.5cm}

This appendix provides both theoretical and experimental material and is organized as follows:
Appendix~\ref{sec:useful_res} presents a classical result, allowing us to characterize the 
RKHS of the graph kernels we introduce. Appendix~\ref{sec:add_exp} provides additional
experimental details that are useful to reproduce our results and additional experimental results.
Then, Appendix~\ref{sec:variants} explains how to accelerate the computation of GCKN when 
using walks instead of paths (at the cost of lower expressiveness), and Appendix~\ref{sec:proofs}
presents a proof of Theorem~\ref{thm:subtree_walk} on the expressiveness of WL and walk kernels.

\section{Useful Result about RKHSs}\label{sec:useful_res}
The following result characterizes the RKHS of a kernel function when an
explicit mapping to a Hilbert space is available. It may be found in classical
textbooks \citep[see, \eg,~][\S 2.1]{saitoh1997integral}.
\begin{theorem}
\label{thm:rkhs}
   Let~$\Phi: \Xcal \to \Fcal$ be a mapping from a data space $\Xcal$ to a Hilbert space~$\Fcal$, and let~$K(x, x') := \langle \Phi(x), \psi(x') \rangle_\Fcal$ for~$x, x'$ in $\Xcal$.
   Consider the Hilbert space
\begin{displaymath}
   \Hcal := \left\{ f_z ~;~ z \in \Fcal \right\} \st f_z: x \mapsto \langle z, \Phi(x) \rangle_\Fcal,
\end{displaymath}
endowed with the norm
\begin{displaymath}
   \|f\|_{\Hcal}^2 := \inf_{z \in \Fcal} ~\left\{ \|z\|_\Fcal^2  \st f  = f_{z} \right\}.
\end{displaymath}
Then, $\Hcal$ is the reproducing kernel Hilbert space associated to kernel~$K$.
\end{theorem}

 \section{Details on Experimental Setup and Additional Experiments}\label{sec:add_exp}
In this section, we provide additional details and more experimental results.
In Section~\ref{subsec:setup}, we provide additional experimental details; 
in Section~\ref{subsec:exps}, we present a benchmark on graph classification 
with continuous attributes by using the protocol of~\citet{togninalli2019wasserstein};
in Section~\ref{subsec:param}, we perform a hyperparameter study for
unsupervised GCKN on three datasets, showing that our approach is relatively 
robust to the choice of hyperparameters. In particular, the number of filters
controls the quality of Nystr\"om's kernel approximation: more filters means a
better approximation and better results, at the cost of more computation. 
This is in contrast with a traditional (supervised) GNN, where more filters may
lead to overfitting. Finally, Section~\ref{subsec:motifs} provides
motif discovery results.

\subsection{Experimental Setup and Reproducibility}\label{subsec:setup}
\paragraph{Hyperparameter search grids.}
In our experiments for supervised models, we use an Adam optimizer~\citep{kingma2014adam} for at most 350 epochs with an initial learning rate equal to 0.01 and halved every 50 epochs with a batch size fixed to 32 throughout all datasets; the number of epochs is selected using cross validation following~\citet{xu2018how}. The full hyperparameter search range is given in Table~\ref{table:hyperparameter} for both unsupervised and supervised models on all tasks. Note that we include some large values (1.5 and 2.0) for $\sigma$ to simulate the linear kernel as we discussed in Section~\ref{sec:multilayer}. In fact, the function $\sigma_1(x)=e^{\alpha(x-1)}$ defined in \eqref{eq:sigma0} is upper bounded by $e^{-\alpha}+(1-e^{-\alpha})x$ and lower bounded by $1+\alpha(x-1)$ by its convexity at $0$ and $1$. Their difference is increasing with $\alpha$ and converges to zero when $\alpha$ tends to $0$. Hence, when $\alpha$ is small, $\sigma_1$ behaves as an affine kernel with a small slope.
\begin{table}[h]
\caption{Hyperparameter search range}\label{table:hyperparameter}
\centering
	\begin{tabular}{lc}
		\toprule
		Hyperparameter & Search range \\ \midrule
		$\sigma$ ($\alpha=1/\sigma^2$) & $[0.3;0.4;0.5;0.6;1.0;1.5;2.0]$ \\
		local/global pooling & [sum, mean, max] \\
		path size $k_1$ & integers between 2 and 12 \\
		number of filters (unsup) & [32; 128; 512; 1024] \\
		number of filters (sup) & [32; 64] and 256 for ENZYMES \\
		$\lambda$ (unsup) & $1/n \times \text{np.logspace(-3, 4, 60)}$ \\
		$\lambda$ (sup)  & [0.01; 0.001; 0.0001; 1e-05; 1e-06; 1e-07] \\
		\bottomrule
	\end{tabular}
\end{table}
\paragraph{Computing infrastructure.}
Experiments for unsupervised models were conducted by using a shared CPU cluster composed of 2 Intel Xeon E5-2470v2 @2.4GHz CPUs with 16 cores and 192GB of RAM.
Supervised models were trained by using a shared GPU cluster, in large parts built with Nvidia gamer cards (Titan X, GTX1080TI). About 20 of these CPUs and 10 of these GPUs were used simultaneously to perform the experiments of this paper.

\subsection{Hyperparameter Study}\label{subsec:param}
\begin{figure}[hbtp]
	\centering
	\includegraphics[width=.3\textwidth]{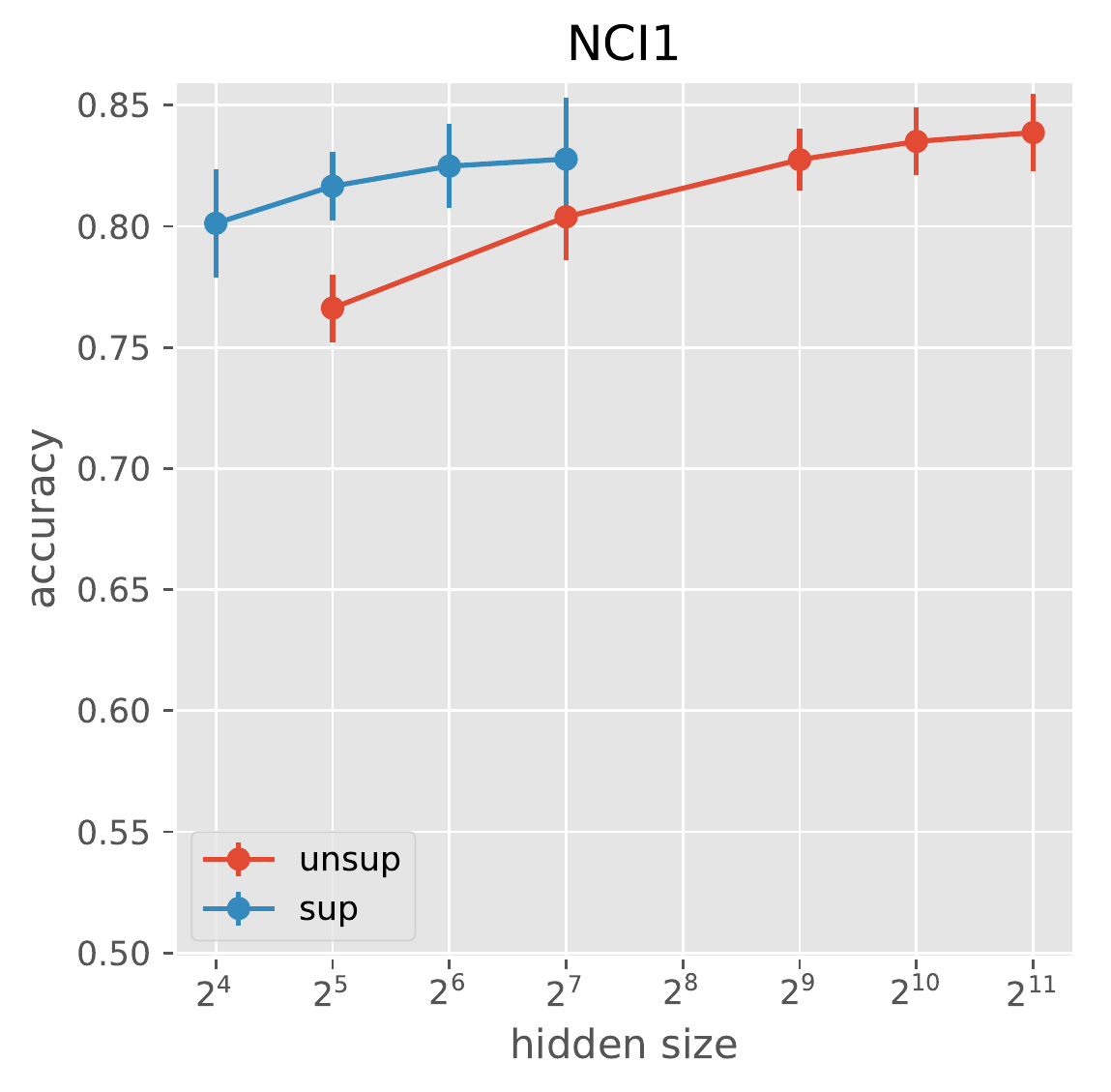}
	\includegraphics[width=.3\textwidth]{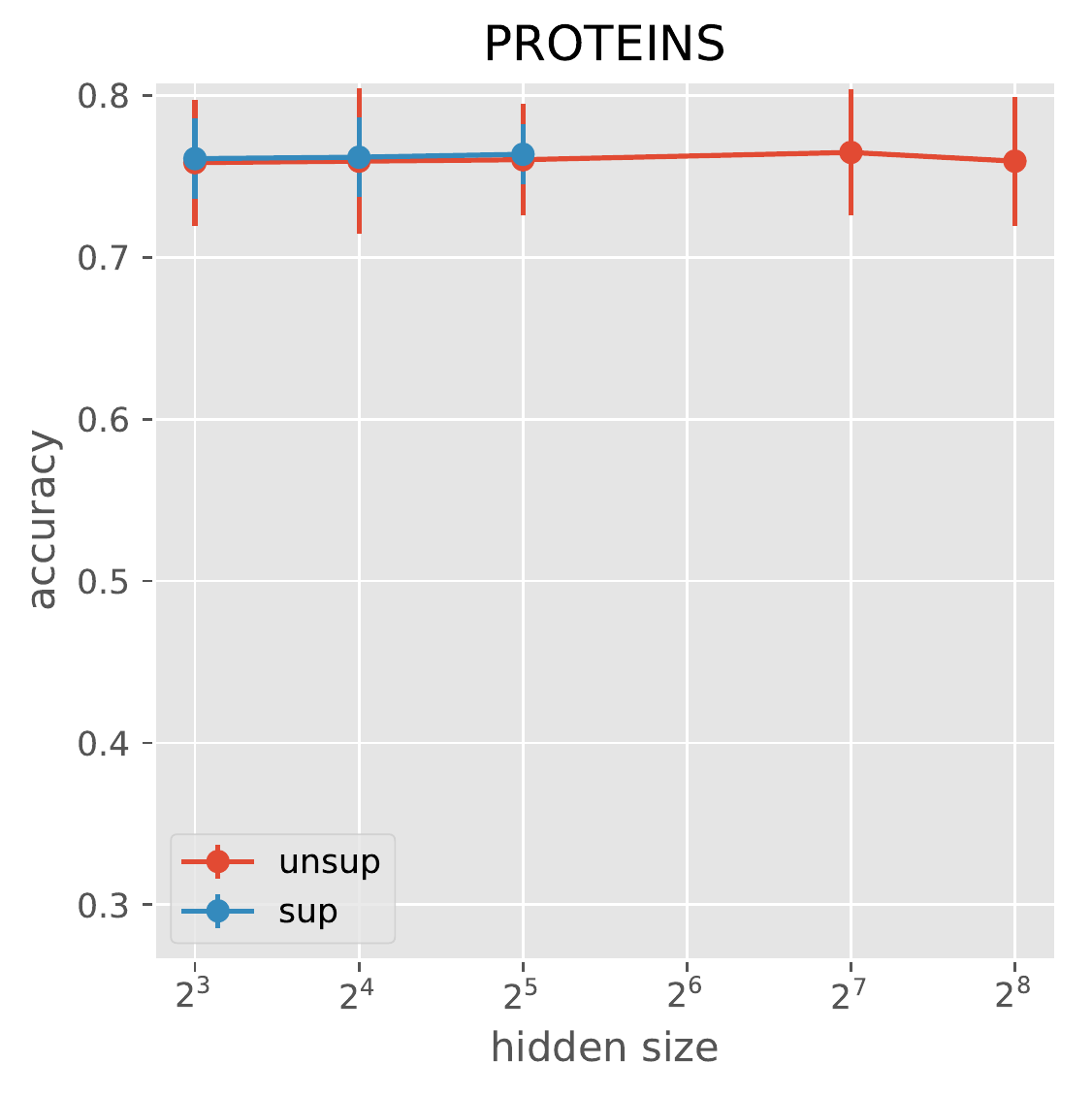}
	\includegraphics[width=.3\textwidth]{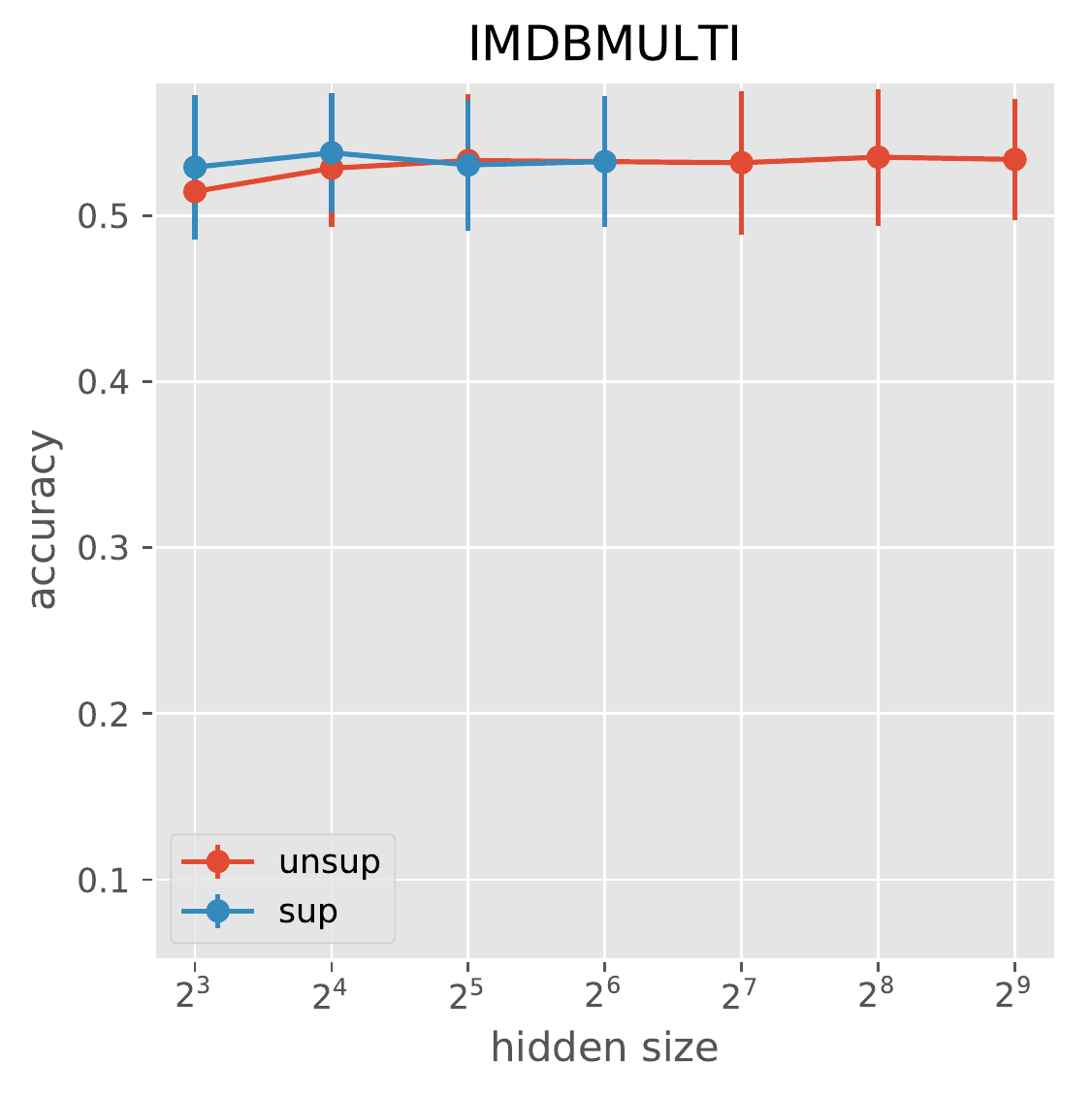} \\
	\includegraphics[width=.3\textwidth]{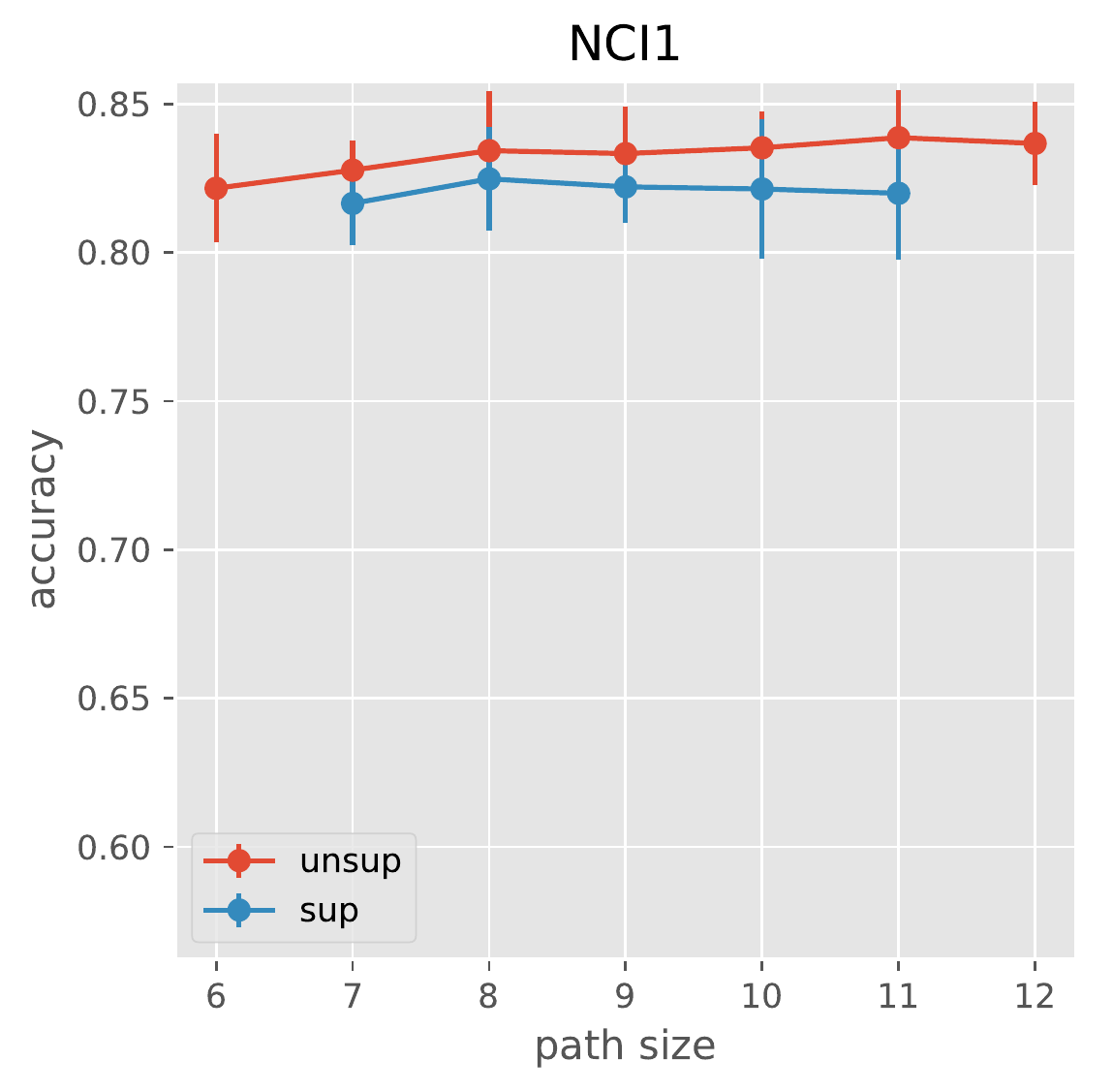}
	\includegraphics[width=.3\textwidth]{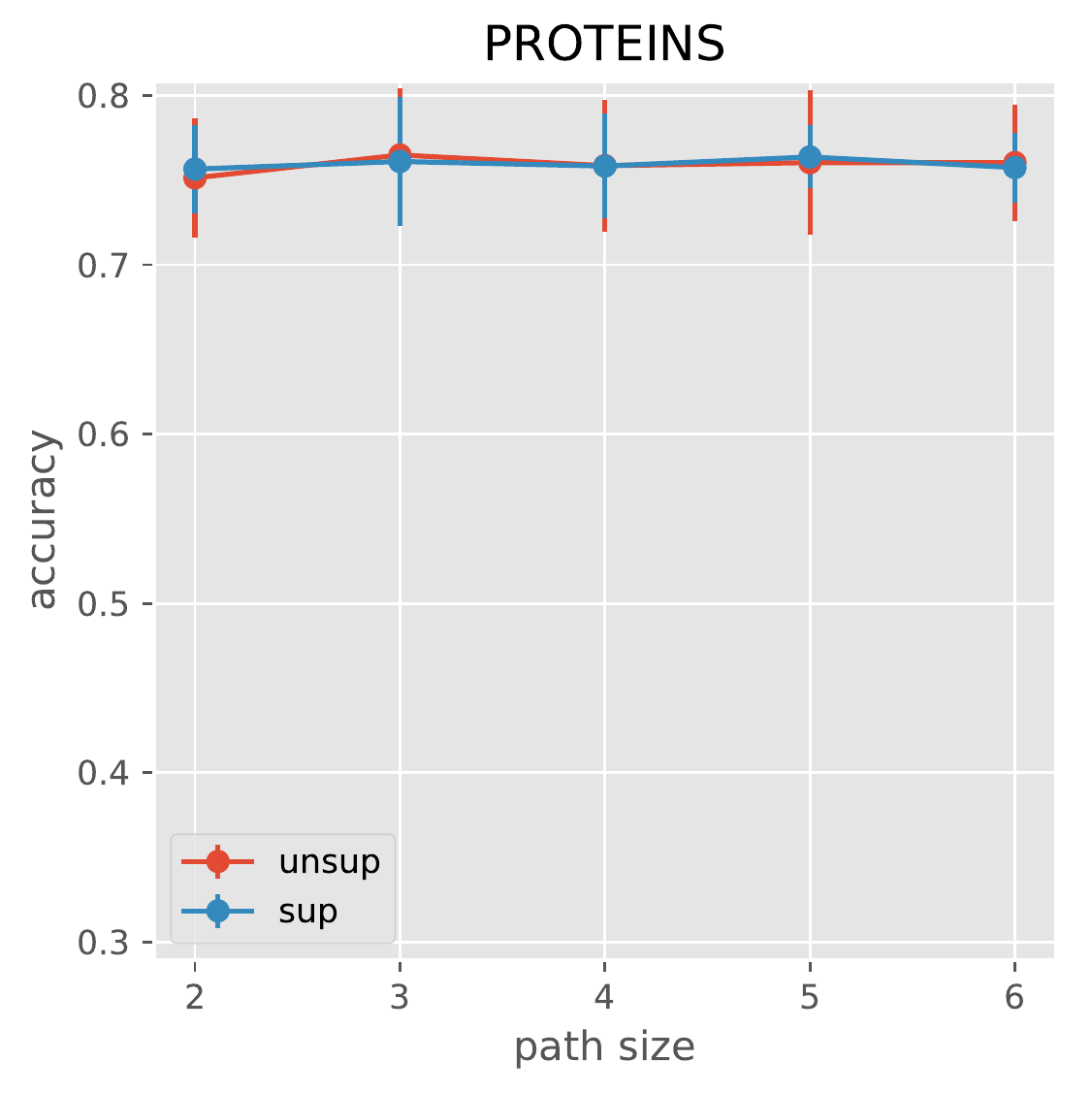}
	\includegraphics[width=.3\textwidth]{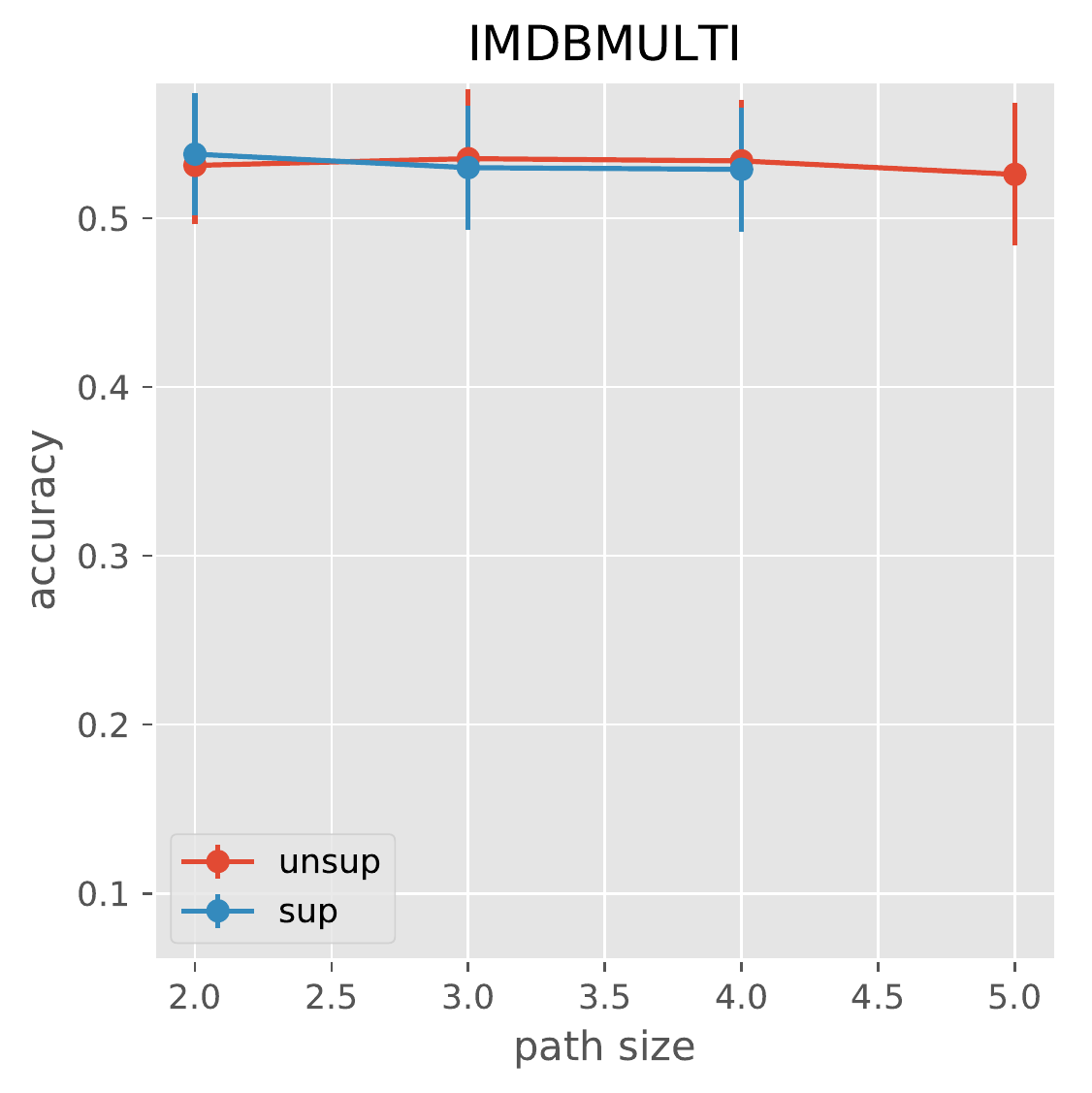} \\
	\includegraphics[width=.3\textwidth]{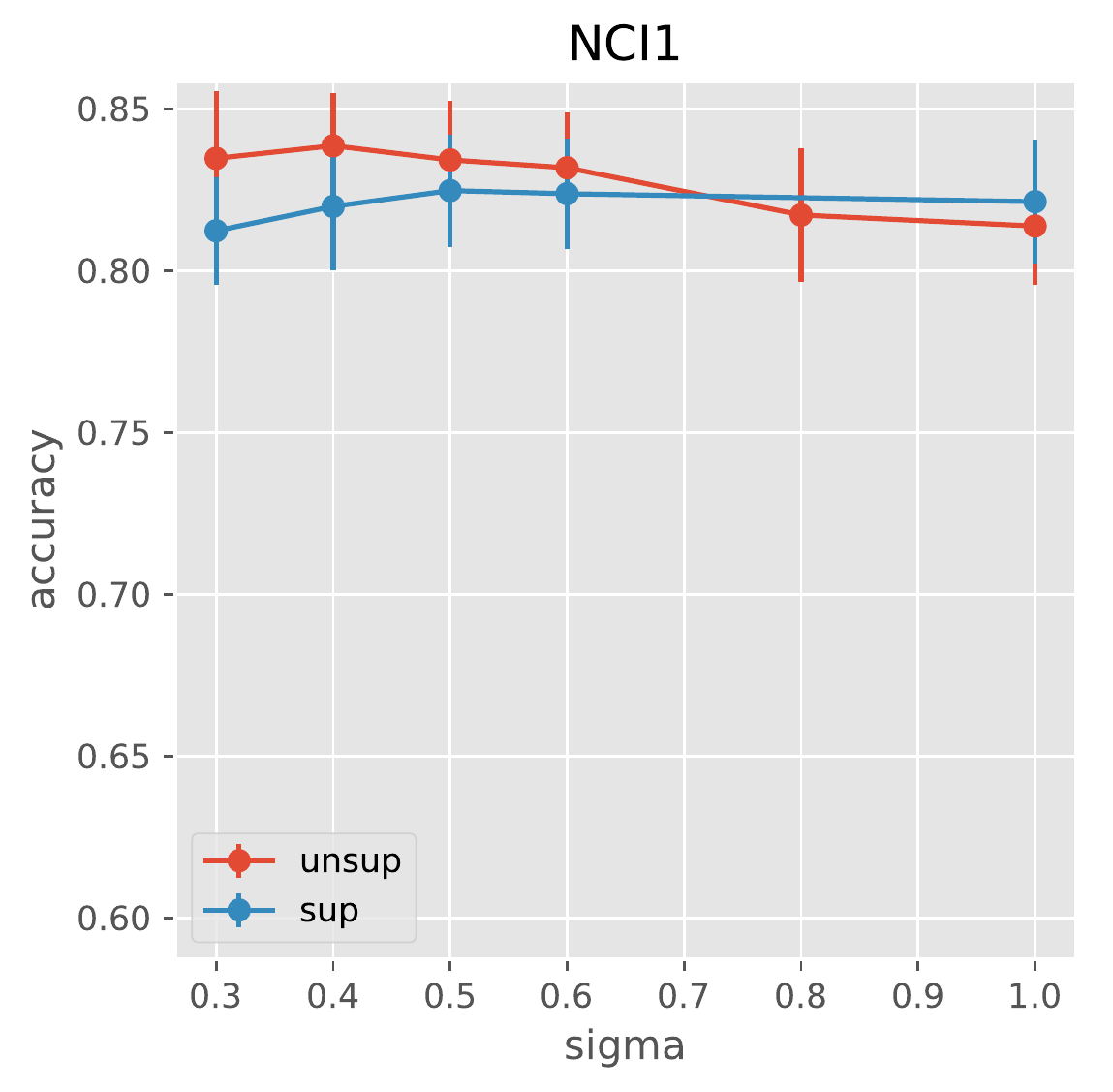}
	\includegraphics[width=.3\textwidth]{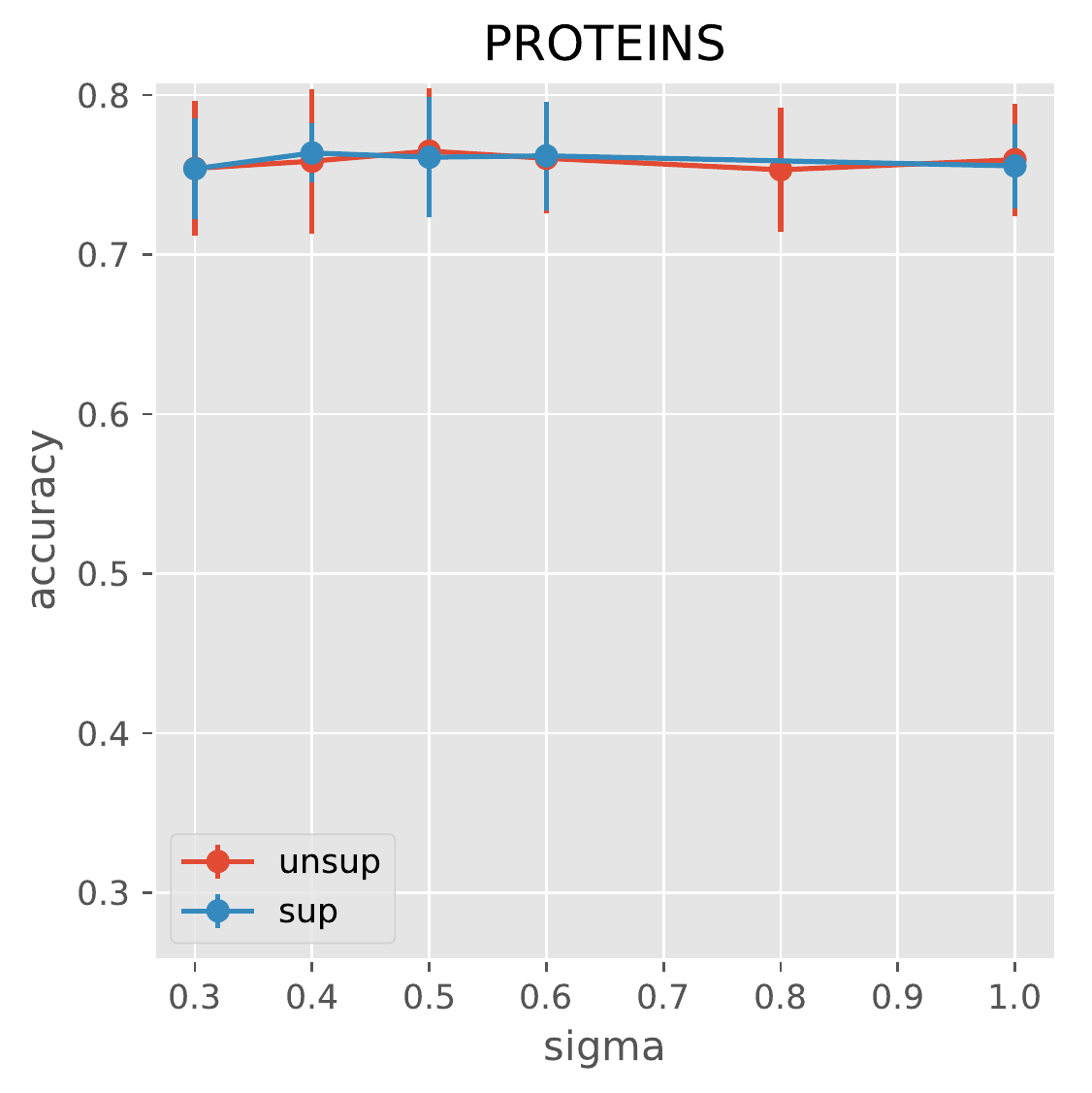}
	\includegraphics[width=.3\textwidth]{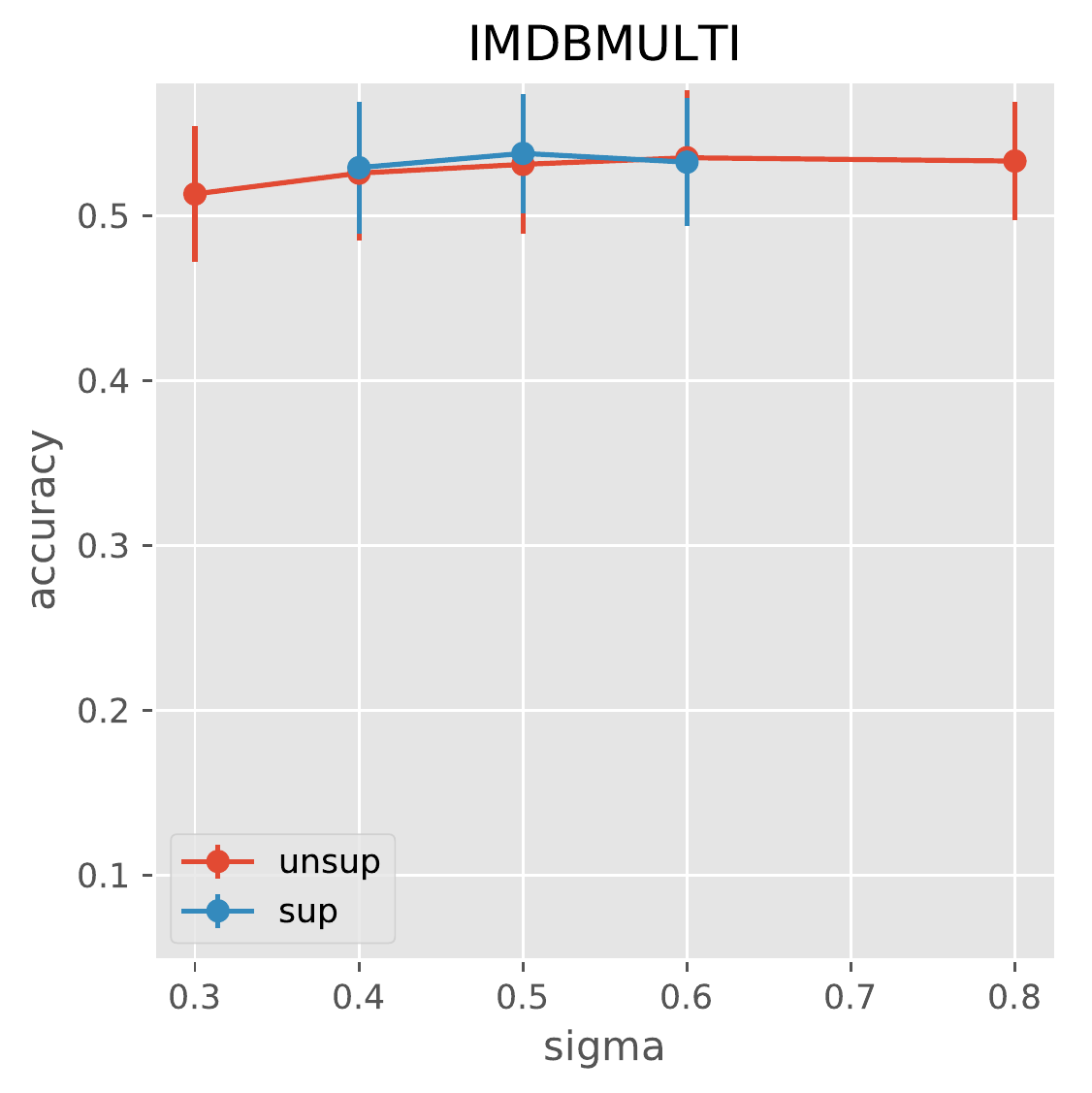} \\
	\includegraphics[width=.3\textwidth]{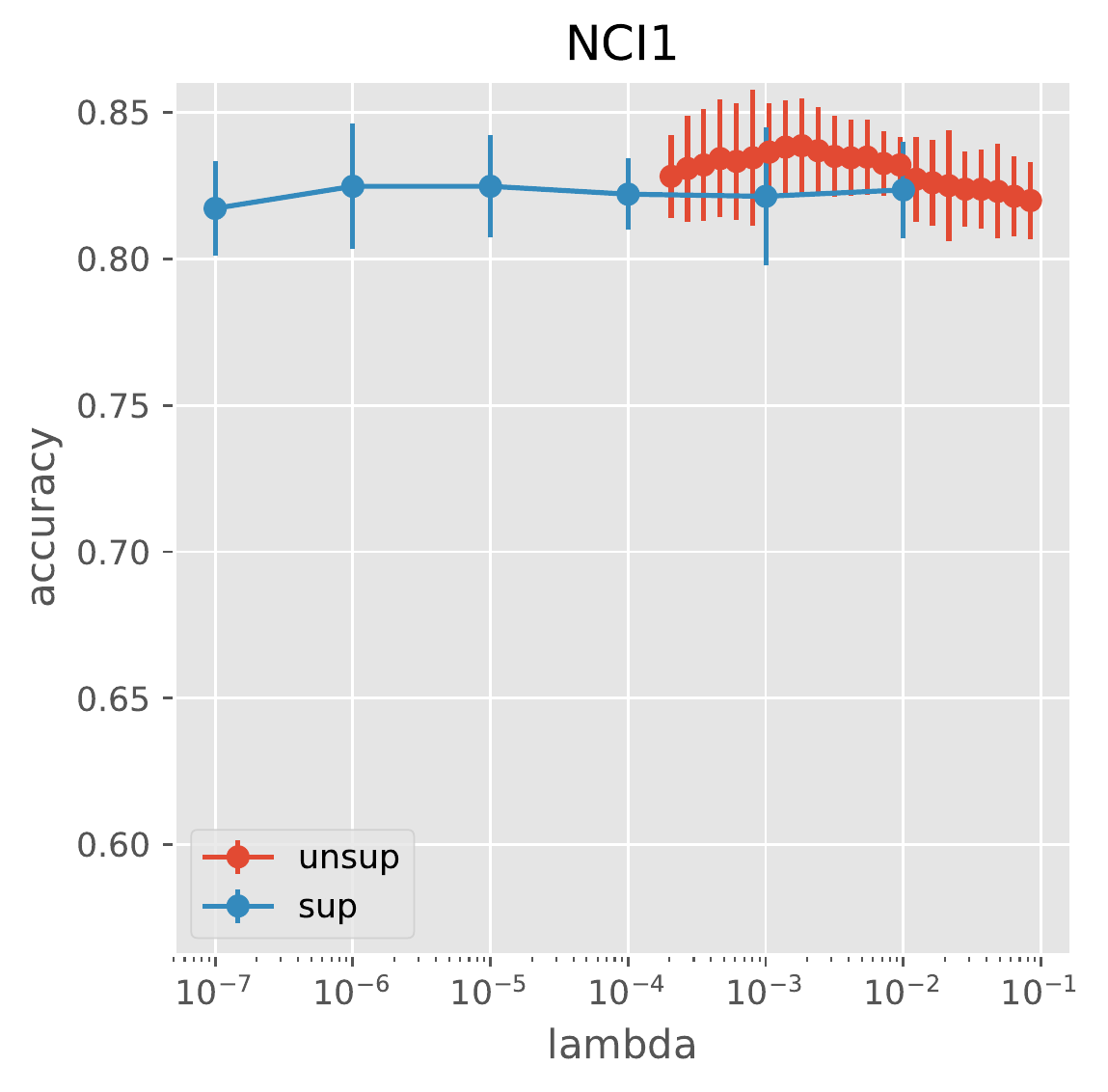}
	\includegraphics[width=.3\textwidth]{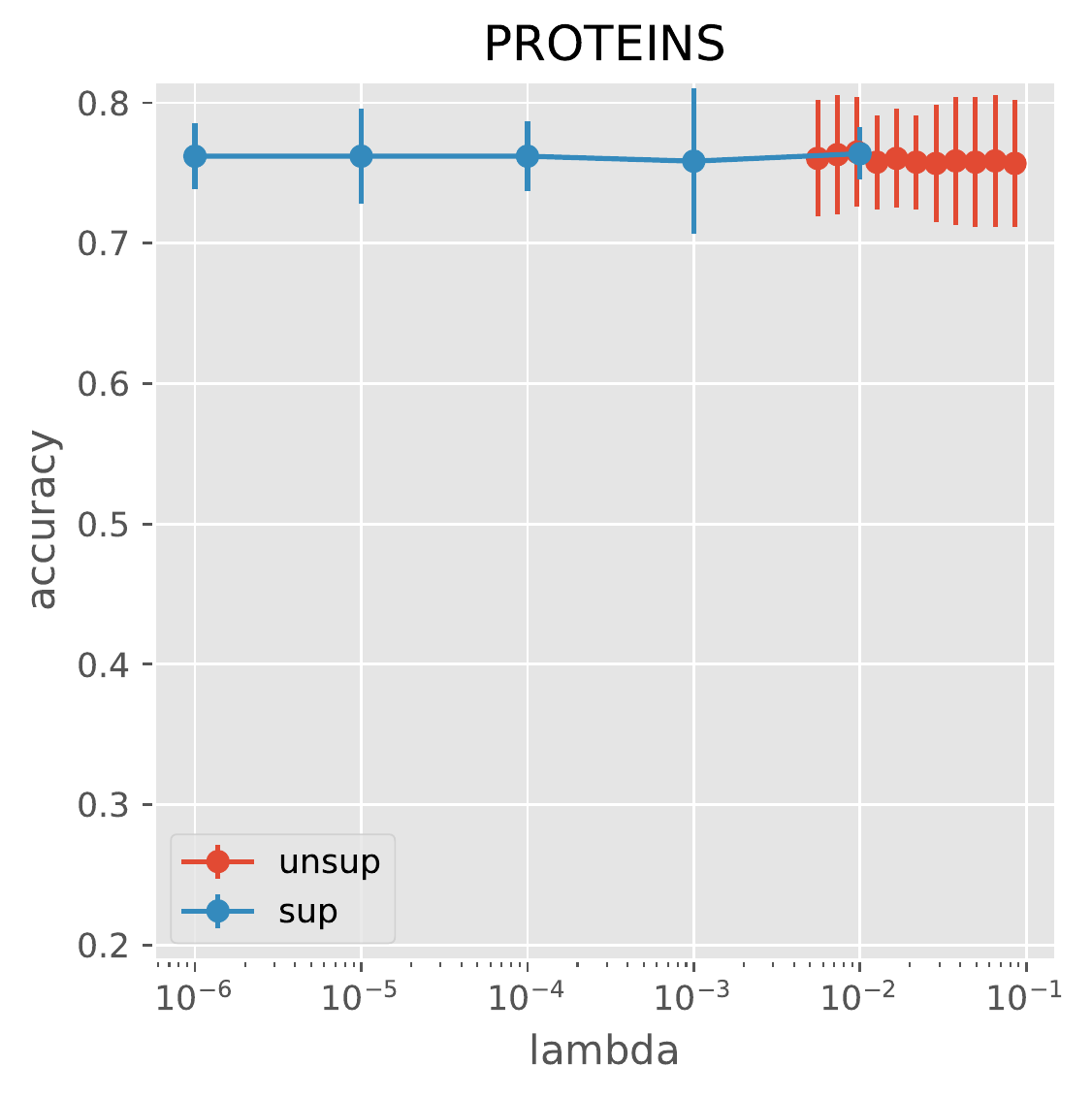}
	\includegraphics[width=.3\textwidth]{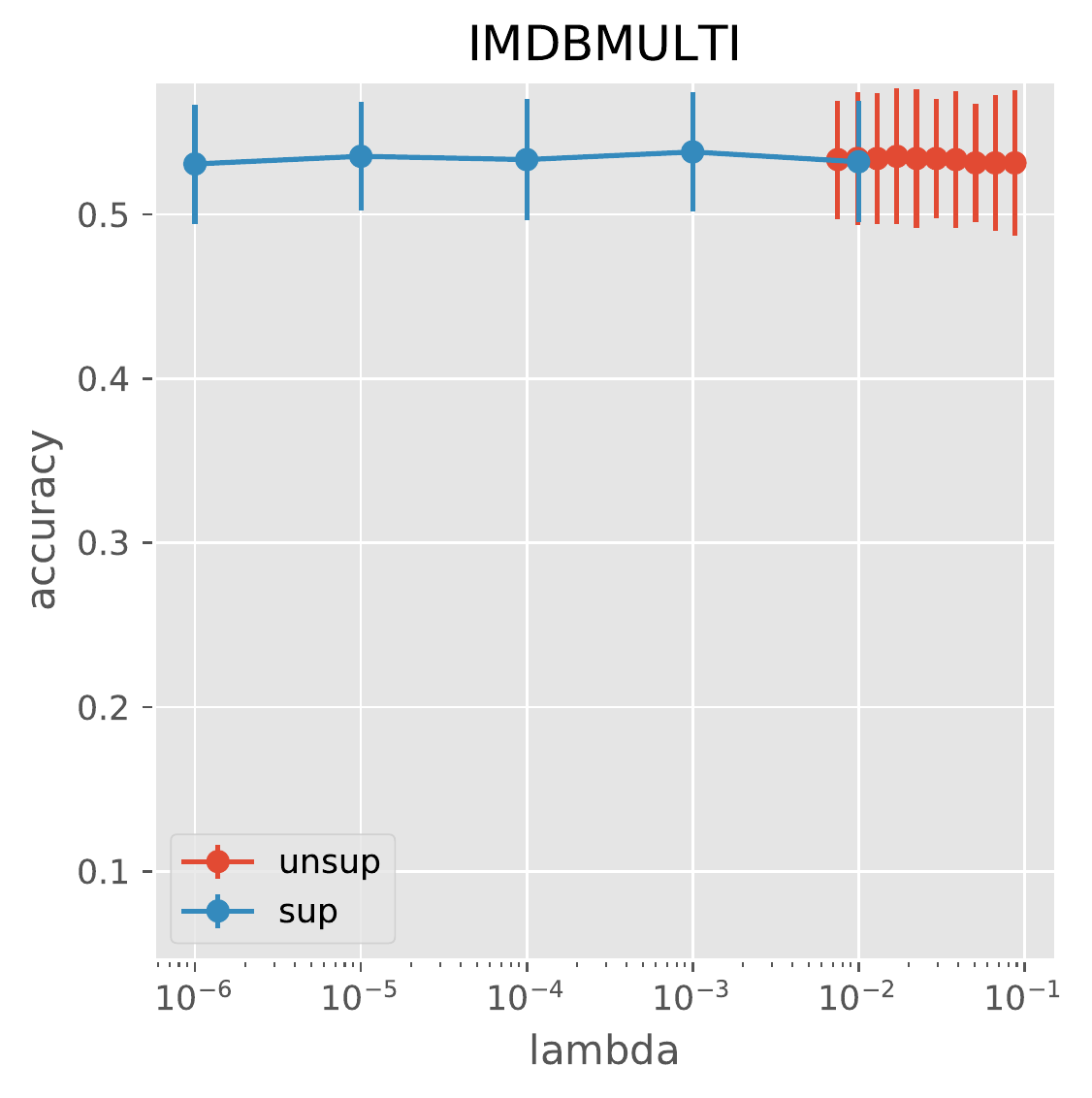} \\
	\caption{Hyperparamter study: sensibility to different hyperparameters for unsupervised and supervised GCKN-subtree models. The row from top to bottom respectively corresponds to number of filters $q_1$, path size $k_1$, bandwidth parameter $\sigma$ and regularization parameter $\lambda$. The column from left to right corresponds to different datasets: NC11, PROTEINS and IMDBMULTI.}\label{fig:hyper_app}
\end{figure}

We show here that both unsupervised and supervised models are generally robust to different hyperparameters, including path size $k_1$, bandwidth parameter $\sigma$, regularization parameter $\lambda$ and their performance grows increasingly with the number of filters $q$. The accuracies for NCI1, PROTEINS and IMDBMULTI are given in Figure~\ref{fig:hyper_app}, by varying respectively the number of filters, the path size, the bandwidth parameter and regularization parameter when fixing other parameters which give the best accuracy. Supervised models generally require fewer number of filters to achieve similar performance to its unsupervised counterpart. In particular on the NCI1 dataset, the supervised GCKN outperforms its unsupervised counterpart by a significant margin when using a small number of filters.

\subsection{Model Interpretation}\label{subsec:motifs}
\begin{figure}
	\centering
	\includegraphics[width=0.4\linewidth]{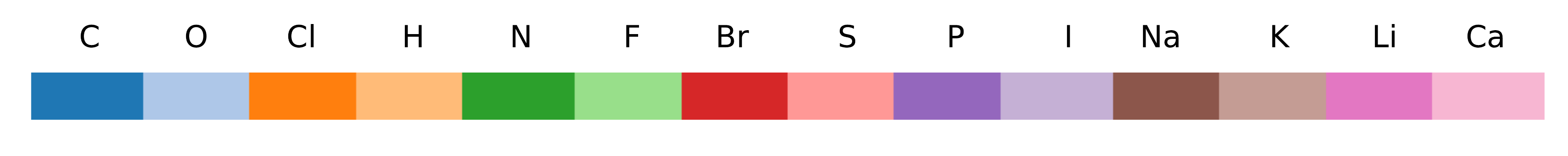} \\
	\includegraphics[width=0.24\linewidth]{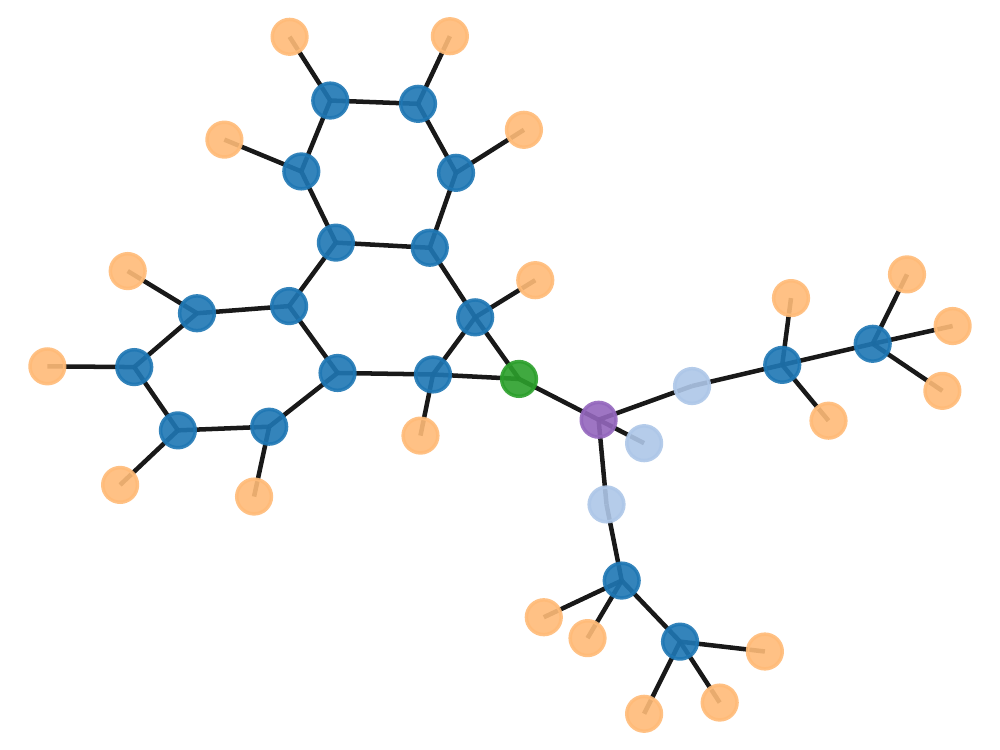}
	\includegraphics[width=0.24\linewidth]{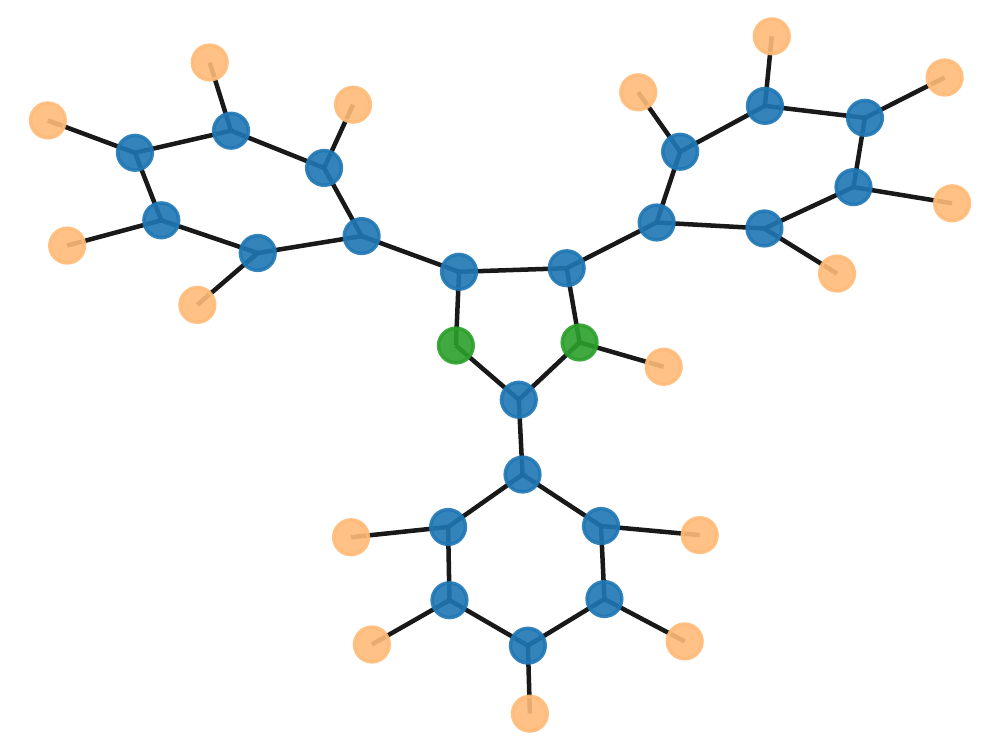}
	\includegraphics[width=0.24\linewidth]{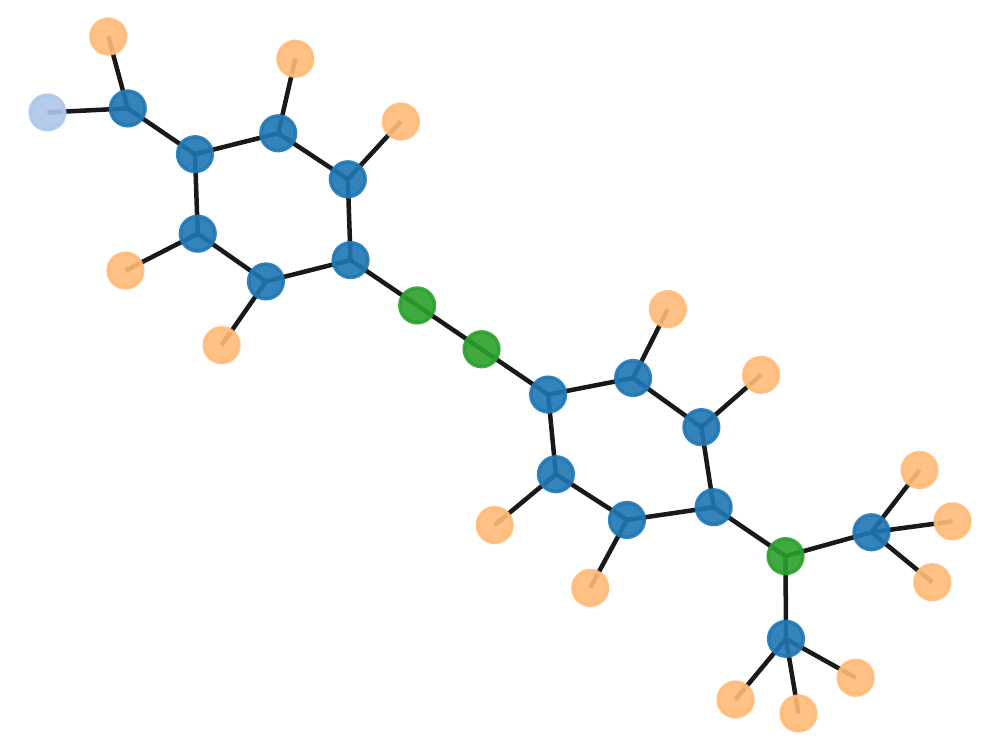}
	\includegraphics[width=0.24\linewidth]{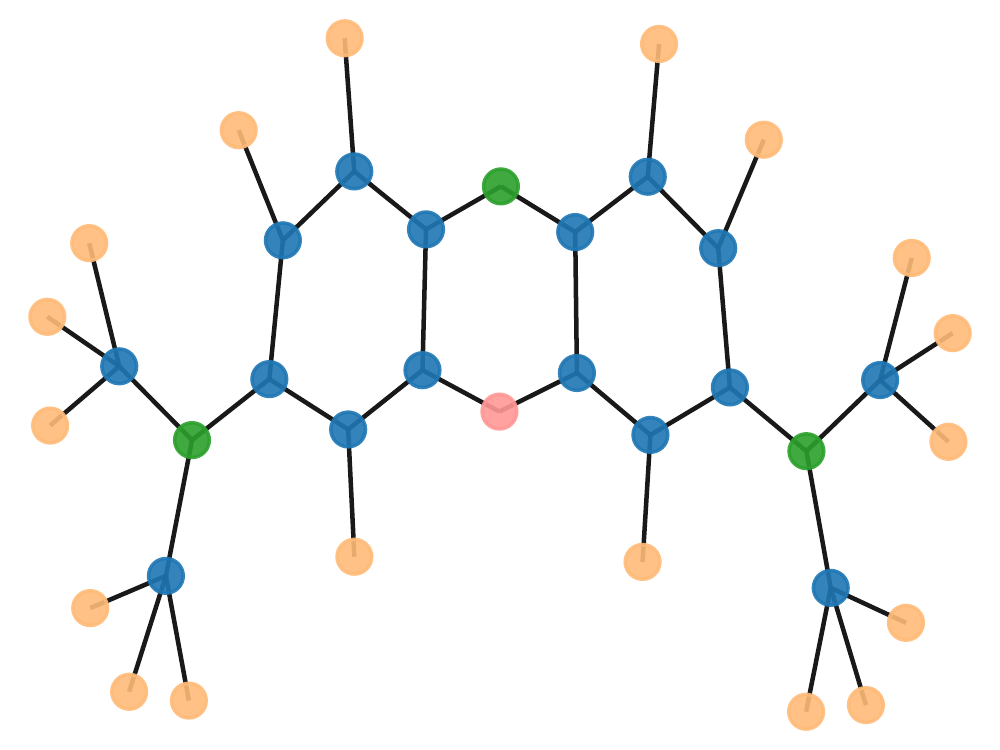}\\
	\includegraphics[width=0.24\linewidth]{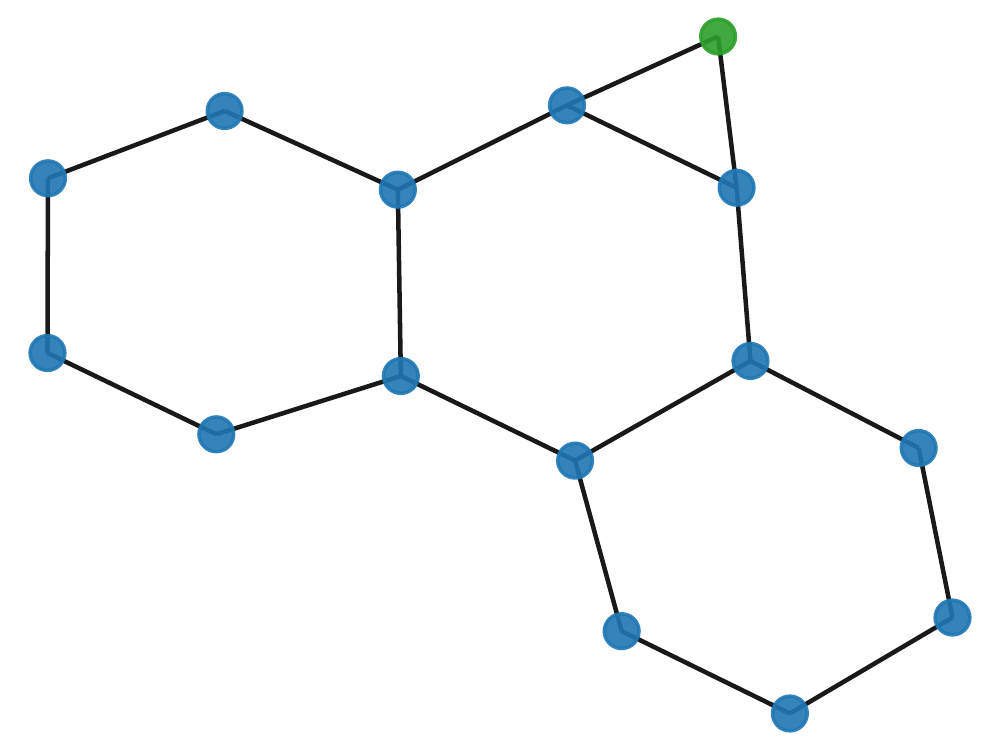}
	\includegraphics[width=0.24\linewidth]{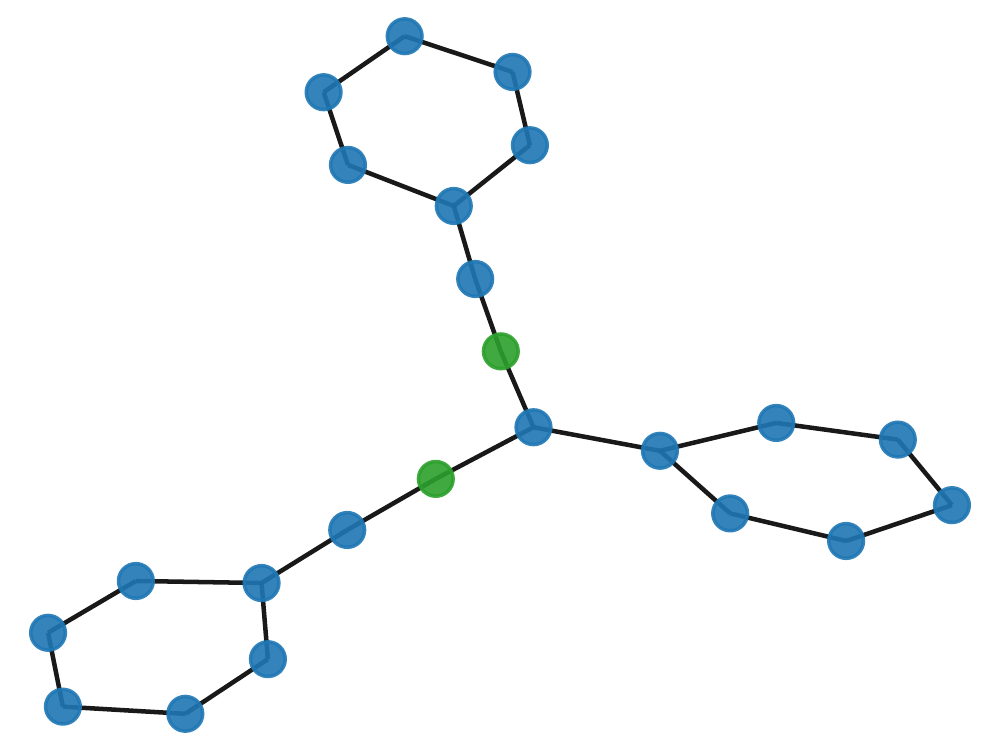}
	\includegraphics[width=0.24\linewidth]{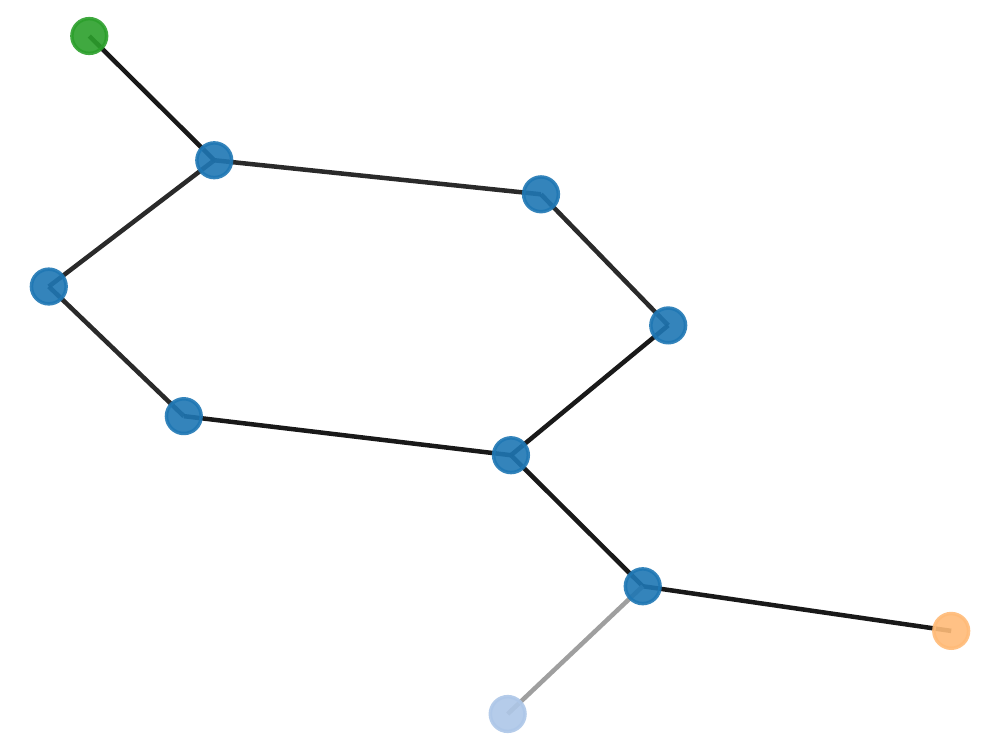}
	\includegraphics[width=0.24\linewidth]{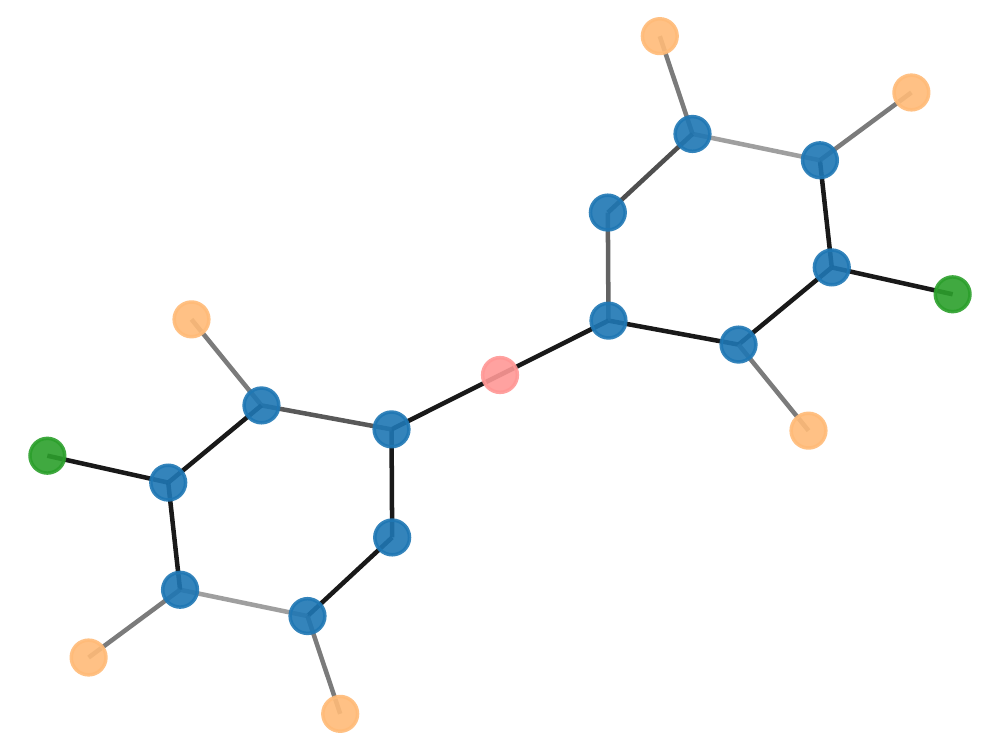} \\ \vspace{1cm}
	\includegraphics[width=0.24\linewidth]{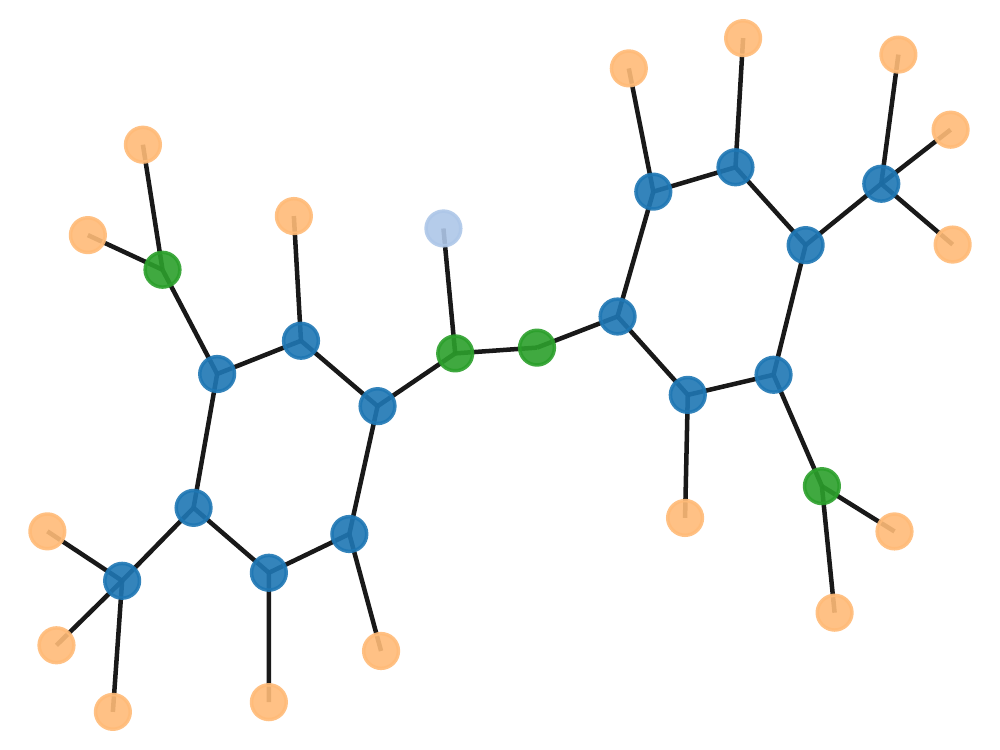}
	\includegraphics[width=0.24\linewidth]{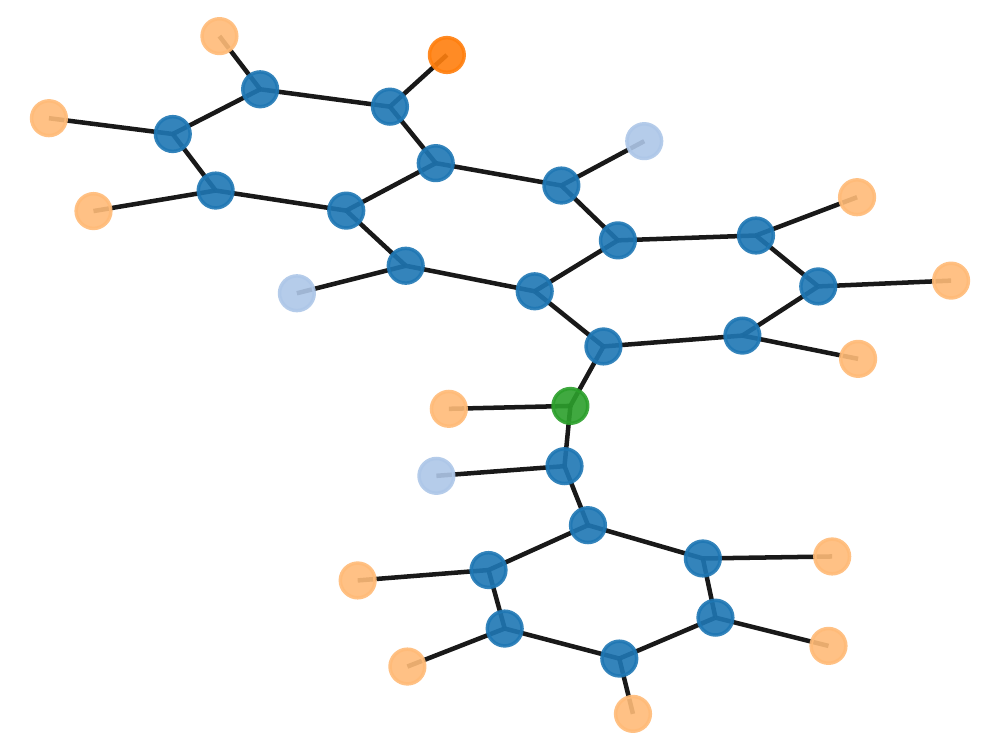}
	\includegraphics[width=0.24\linewidth]{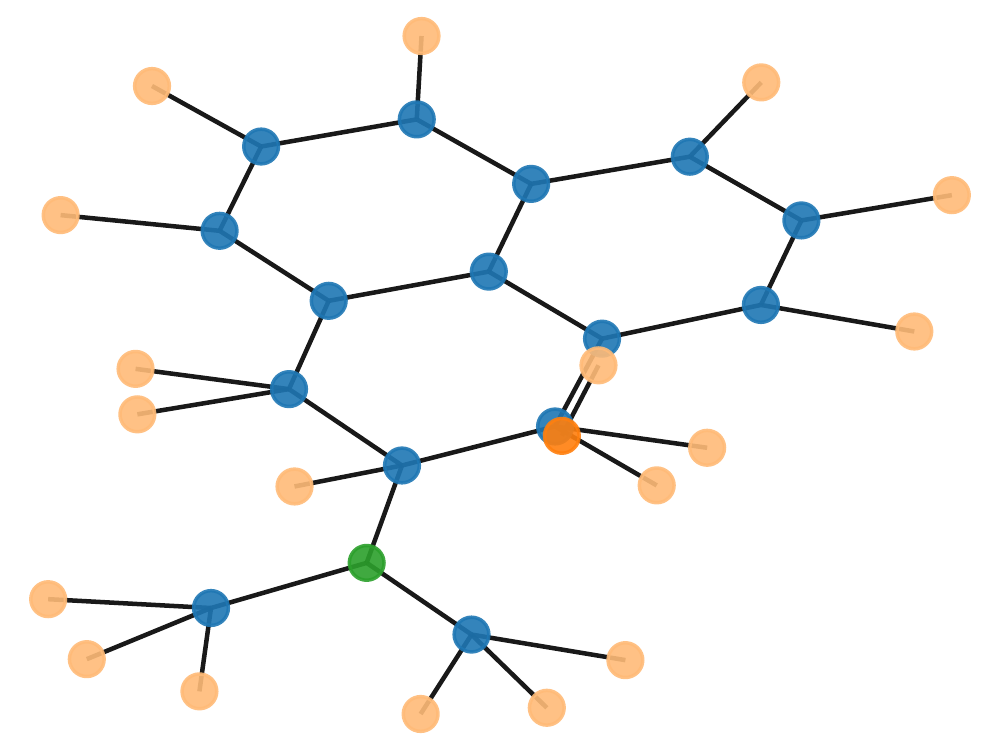}
	\includegraphics[width=0.24\linewidth]{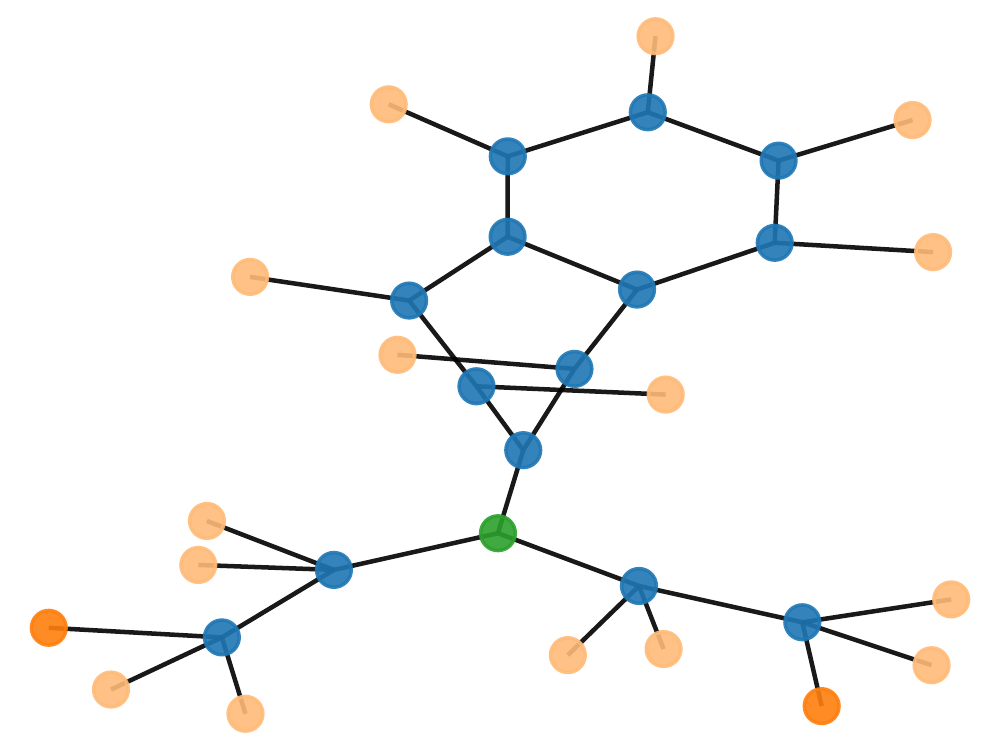}\\
	\includegraphics[width=0.24\linewidth]{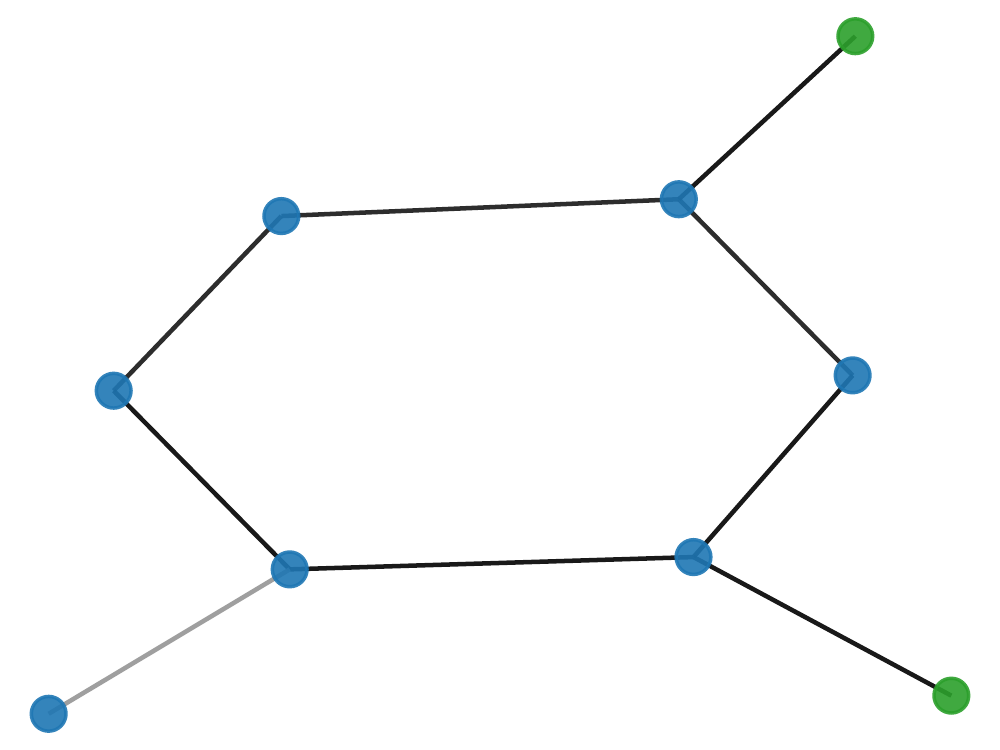}
	\includegraphics[width=0.24\linewidth]{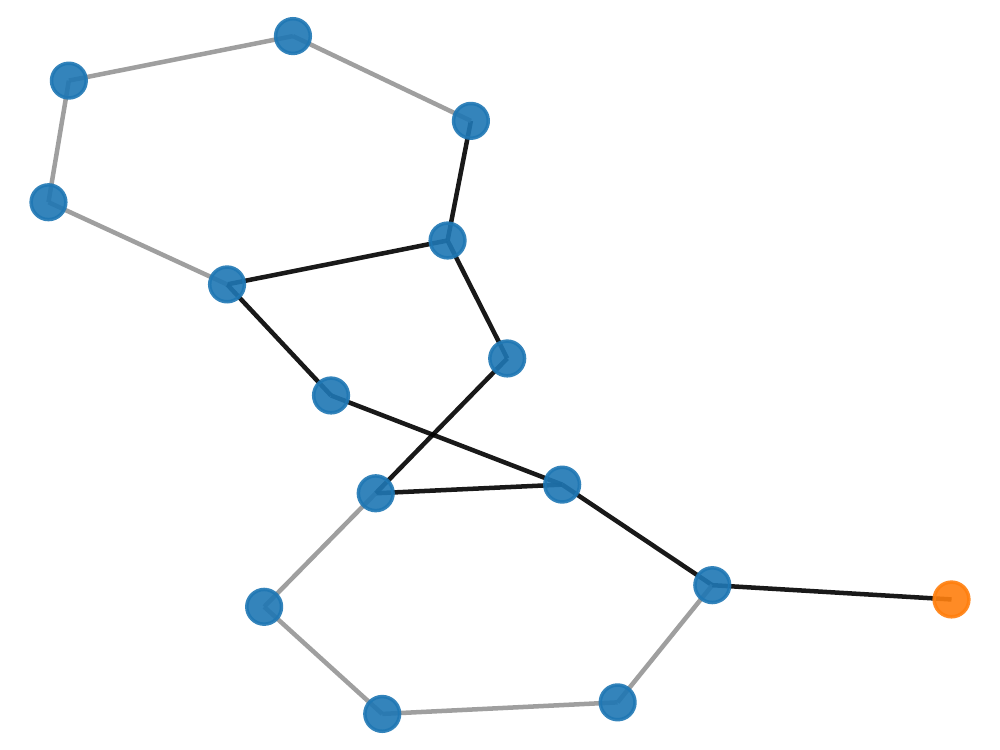}
	\includegraphics[width=0.24\linewidth]{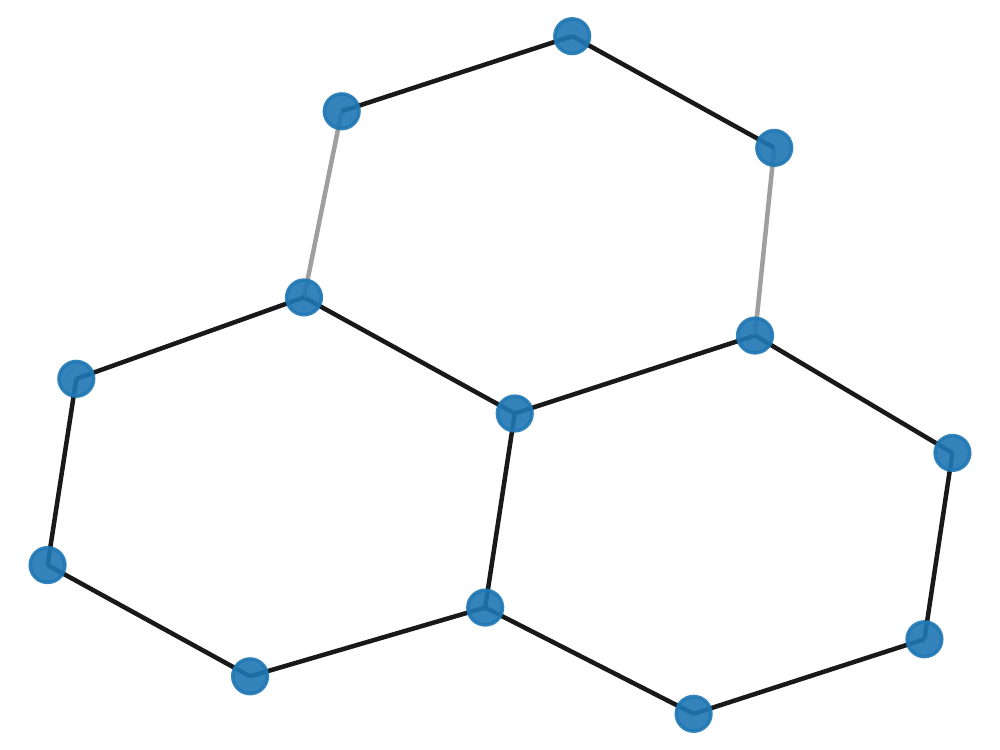}
	\includegraphics[width=0.24\linewidth]{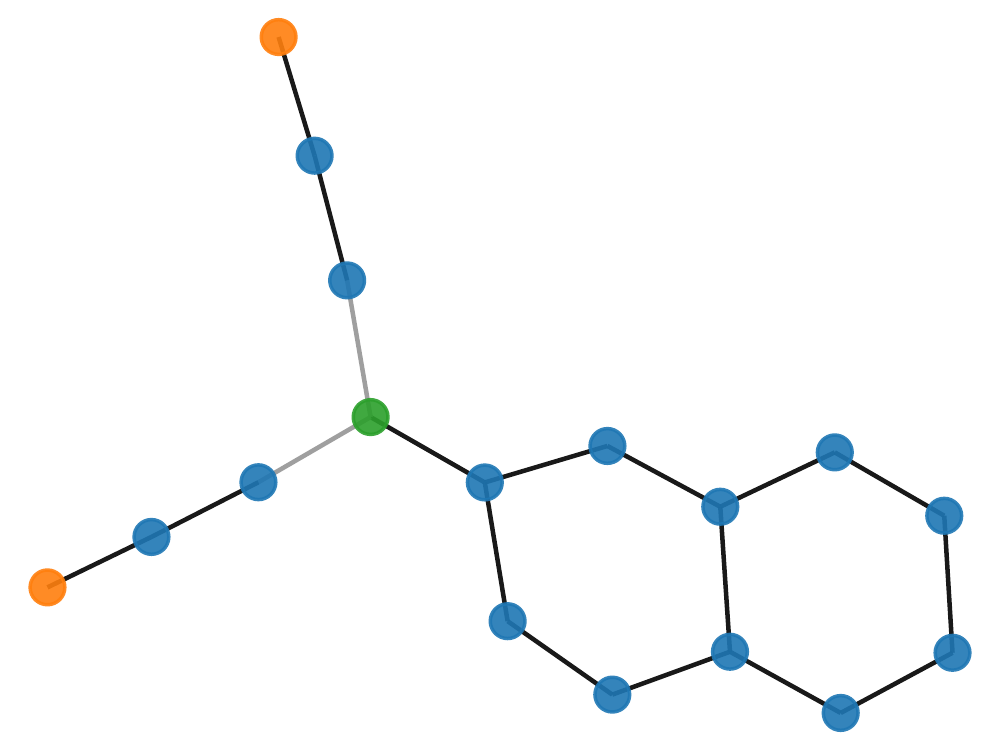}
	\caption{More motifs extracted by GCKN on Mutagenicity dataset. First and third rows are original graphs; second and fourth rows are corresponding motifs. Some benzene ring or polycyclic aromatic groups are identified, which are known to be mutagenic. In addition, Some chemical groups whose mutagenicity is not known are also identified, such as polyphenylene sulfide in the fourth subgraph and 2-chloroethyl- in the last subgraph.}\label{fig:motif_app}
\end{figure}

\paragraph{Implementation details.}
We use a similar experimental setting as \citet{ying2019gnnexplainer} to train a supervised GCKN-subtree model on Mutagenicity dataset, consisting of 4337 molecule graphs labeled according to their mutagenic effect. Specifically, we use the same split for train and validation set and train a GCKN-subtree model with $k_1=3$, which is similar to a 3-layer GNN model. The number of filters is fixed to 20, the same as \citet{ying2019gnnexplainer}. The bandwidth parameter $\sigma$ is fixed to 0.4, local and global pooling are fixed to mean pooling, the regularization parameter $\lambda$ is fixed to 1e-05. We use an Adam optimizer with initial learning equal to 0.01 and halved every 50 epochs, the same as previously. The accuracy of the trained model is assured to be more than 80\% on the test set as \citet{ying2019gnnexplainer}. Then we use the procedure described in Section~\ref{sec:interpretation} to interpret our trained model. We use an LBFGS optimizer and fixed $\mu$ to 0.01. The final subgraph for each given graph is obtained by extracting the maximal connected component formed by the selected paths. A contribution score for each edge can also be obtained by gathering the weights $M$ of all the selected paths that pass through this edge.
\paragraph{More results.}
More motifs extracted by GCKN are shown in Figure~\ref{fig:motif_app} for the Mutagenicity dataset. We recovered some benzene ring or polycyclic aromatic groups which are known to be mutagenic. We also found some groups whose mutagenicity is not known, such as polyphenylene sulfide in the fourth subgraph and 2-chloroethyl- in the last subgraph.

 \section{Fast Computation of GCKN with Walks}\label{sec:variants}
Here we discuss an efficient computational variant using walk kernel instead of path kernel, at the cost of losing some expressive power.
Let us consider a relaxed walk kernel by analogy to~\eqref{eq:kbase} with
\begin{equation}
   \kappa_{\text{base}}^{(k)}(u,u') =   \sum_{p\in\Wcal_k(G,u)}
   \sum_{p'\in\Wcal_k(G',u')} \kappa_1(\varphi_0(p),\varphi_0'(p')),   
\end{equation}
using walks instead of paths and with $\kappa_1$ the Gaussian kernel defined in~\eqref{eq:gaussian}. As Gaussian kernel can be decomposed as a product of the Gaussian kernel on pair of nodes at each position
\begin{equation*}
	\kappa_1(\varphi_0(p),\varphi_0'(p') )=\prod_{j=1}^{k} \kappa_1(\varphi_0(p_j),\varphi_0'(p'_j) ),
\end{equation*}
We can obtain similar recursive relation as for the original walk kernel in Lemma~\ref{lm:walk}
\begin{equation}\label{eq:rec_walk}
	\kappa_{\text{base}}^{(k)}(u,u')=\kappa_1(\varphi_0(u),\varphi_0'(u') )\sum_{v\in\Ncal(u)} \sum_{v'\in\Ncal(u')} \kappa_{\text{base}}^{(k-1)}(v, v').
\end{equation}
After applying the Nystr\"om method, the approximate feature map in~\eqref{eq:psi} becomes
\begin{equation*}
	\psi_1(u)=\sigma_1(Z^{\top} Z)^{-\frac{1}{2}} c_k(u),
\end{equation*}
where for any $0 \leq j\leq k$, $c_j(u):=\sum_{p\in\Wcal_j(G,u)} \sigma_1(Z^{\top}_j \psi_0(p) )$ and $Z_j$ in $\Real^{q_0 (j+1)\times q_1}$ denotes the matrix consisting of the $j+1$ last columns of $q_1$ anchor points. Using the above recursive relation \eqref{eq:rec_walk} and similar arguments in \eg~\cite{chen2019rec}, we can show $c_j$ obeys the following recursive relation
\begin{equation}
	c_j(u)=b_j(u)\odot \sum_{v\in\Ncal(u)} c_{j-1}(v),~ 1\leq j\leq k,
\end{equation}
where $\odot$ denotes the element-wise product and $b_j(u)$ is a vector in $\Real^{q_1}$ whose entry $i$ in $\{1,\dots,q_1\}$ is $\kappa_1(u,z_{i}^{(k+1-j)})$ and $z_{i}^{(k+1-j)}$ denotes the $k+1-j$-th column vector of $z_i$ in $\Real^{q_0}$.
In practice, $\sum_{v\in\Ncal(u)} c_{j-1}(v)$ can be computed efficiently by multiplying the adjacency matrix with the $|\Vcal|$-dimensional vector with entries $c_{j-1}(v)$ for $v\in\Vcal$. \section{Proof of Theorem 1}\label{sec:proofs}

Before presenting and proving the link between the WL subtree kernel and the walk kernel, we start by reminding and showing some useful results about the WL subtree kernel and the walk kernel.
\subsection{Useful results for the WL subtree kernel}
We first recall a recursive relation of the WL subtree kernel, given in the Theorem 8 of \citet{shervashidze2011weisfeiler}. Let us denote by $\Mcal(u,u')$ the set of exact matchings of subsets of the neighbors of $u$ and $u'$, formally given by
\begin{multline}
		\Mcal(u,u')=\Big\{ R\subseteq \Ncal(u)\times\Ncal(u')\,\Big|\, |R|=|\Ncal(u)|=|\Ncal(u')| \land \\ (\forall (v,v'),(w,w')\in R: u=w \Leftrightarrow u'=w') \land (\forall (u,u')\in R: a(u)=a'(u')) \Big\}.
\end{multline}
Then we have the following recursive relation for $\kappa_{\text{subtree}}^{(k)}(u,u'):=\delta(a_k(u), a_k'(u'))$
\begin{equation}\label{eq:recursion_wl}
	\kappa_{\text{subtree}}^{(k+1)}(u,u')=
	\begin{cases}
		\kappa_{\text{subtree}}^{(k)}(u,u')\max\limits_{R\in\Mcal(u,u')} \prod\limits_{(v, v')\in R} \kappa_{\text{subtree}}^{(k)}(v,v') , & \text{if } \Mcal(u,u')\neq \emptyset,\\
		0, & \text{otherwise.}
	\end{cases}
\end{equation}
We can further simply the above recursion using the following Lemma
\begin{lemma}\label{lm:wl_rec}
	If $\Mcal(u,u')\neq \emptyset$, we have
	\begin{equation*}
		\kappa_{\text{subtree}}^{(k+1)}(u,u')=\delta(a(u),a'(u')) \max_{R\in\Mcal(u,u')} \prod_{(v, v')\in R} \kappa_{\text{subtree}}^{(k)}(v,v').
	\end{equation*}
\end{lemma}
\begin{proof}
	We prove this by induction on $k\geq 0$. For $k=0$, this is true by the definition of $\kappa_{\text{subtree}}^{(0)}$. For $k\geq 1$, we suppose that $\kappa_{\text{subtree}}^{(k)}(u,u')=\delta(a(u),a'(u')) \max_{R\in\Mcal(u,u')} \prod_{(v, v')\in R} \kappa_{\text{subtree}}^{(k -1)}(v,v')$. We have
	\begin{equation*}
		\begin{aligned}
			\kappa_{\text{subtree}}^{(k+1)}(u,u')&=\kappa_{\text{subtree}}^{(k)}(u,u')\max\limits_{R\in\Mcal(u,u')} \prod\limits_{(v, v')\in R} \kappa_{\text{subtree}}^{(k)}(v,v') \\
			&=\delta(a(u),a'(u')) \max_{R\in\Mcal(u,u')} \prod_{(v, v')\in R} \kappa_{\text{subtree}}^{(k -1)}(v,v') \max\limits_{R\in\Mcal(u,u')} \prod\limits_{(v, v')\in R} \kappa_{\text{subtree}}^{(k)}(v,v').
		\end{aligned}
	\end{equation*}
	It suffices to show
	\begin{equation*}
		\max_{R\in\Mcal(u,u')} \prod_{(v, v')\in R} \kappa_{\text{subtree}}^{(k -1)}(v,v') \max\limits_{R\in\Mcal(u,u')} \prod\limits_{(v, v')\in R} \kappa_{\text{subtree}}^{(k)}(v,v')=\max\limits_{R\in\Mcal(u,u')} \prod\limits_{(v, v')\in R} \kappa_{\text{subtree}}^{(k)}(v,v').
	\end{equation*}
	Since the only values can take for $\kappa_{\text{subtree}}^{(k -1)}$ is 0 and 1, the only values that $\max_{R\in\Mcal(u,u')} \prod_{(v, v')\in R} \kappa_{\text{subtree}}^{(k -1)}(v,v')$ can take is also 0 and 1. Then we can split the proof on these two conditions. It is obvious if this term is equal to 1. If this term is equal to 0, then
	\begin{equation*}
		\max\limits_{R\in\Mcal(u,u')} \prod\limits_{(v, v')\in R} \kappa_{\text{subtree}}^{(k)}(v,v') \leq \max_{R\in\Mcal(u,u')} \prod_{(v, v')\in R} \kappa_{\text{subtree}}^{(k -1)}(v,v')=0,
	\end{equation*}
	as all terms are not negative and $\kappa_{\text{subtree}}^{(k)}(v,v')$ is not creasing on $k$. Then $\max\limits_{R\in\Mcal(u,u')} \prod\limits_{(v, v')\in R} \kappa_{\text{subtree}}^{(k)}(v,v')=0$ and we have 0 for both sides.
\end{proof}

\subsection{Recursive relation for the walk kernel}
We recall that the $k$-walk kernel is defined as
\begin{equation*}
	K(G, G')=\sum_{u\in \Vcal} \sum_{u'\in \Vcal'} \kappa_{\text{walk}}^{(k)}(u,u'),
\end{equation*}
where
\begin{equation*}
	\kappa_{\text{walk}}^{(k)}(u,u') = \sum_{p\in\Wcal_k(G,u)} \sum_{p'\in\Wcal_k(G',u')} \delta(a(p), a'(p')).
\end{equation*}
The feature map of this kernel is given by
\begin{equation*}
	\varphi_{\text{walk}}^{(k)}(u)=\sum_{p\in\Wcal_k(G,u)} \varphi_{\delta}(a(p)),
\end{equation*}
where $\varphi_{\delta}$ is the feature map associated with $\delta$.
We give here a recursive relation for the walk kernel on the size of walks, thanks to its allowance of nodes to repeat.
\begin{lemma}\label{lm:walk}
	For any $k\geq 0$, we have
	\begin{equation}
		\kappa_{\text{walk}}^{(k+1)}(u, u')=\delta(a(u),a'(u')) \sum_{v\in\Ncal(u)} \sum_{v'\in\Ncal(u')} \kappa_{\text{walk}}^{(k)}(v, v').
	\end{equation}
\end{lemma}
\begin{proof}
Noticing that we can always decompose a path $p\in\Wcal_{k+1}(G,u)$, with $(u,v)$ the first edge that it passes and $v\in\Ncal(u)$, into $(u,q)$ with $q\in\Wcal_k(G,v)$, then we have
	\begin{equation*}
		\begin{aligned}
			\kappa_{\text{walk}}^{(k+1)}(u, u')&=\sum_{p\in\Wcal_{k+1}(G,u)} \sum_{p'\in\Wcal_{k+1}(G',u')} \delta(a(p), a'(p')) \\
			&=\sum_{v\in\Ncal(u)}\sum_{p\in\Wcal_k(G,v)} \sum_{v'\in\Ncal(u')}\sum_{p'\in\Wcal_k(G,v')} \delta(a(u),a'(u'))\delta(a(p),a'(p')) \\
			&=\delta(a(u),a'(u'))\sum_{v\in\Ncal(u)} \sum_{v'\in\Ncal(u')} \sum_{p\in\Wcal_k(G,v)} \sum_{p'\in\Wcal_{k}(G',v')} \delta(a(p), a'(p')) \\
			&=\delta(a(u),a'(u')) \sum_{v\in\Ncal(u)} \sum_{v'\in\Ncal(u')} \kappa_{\text{walk}}^{(k)}(v, v').
		\end{aligned}
	\end{equation*}
\end{proof}
This relation also provides us a recursive relation for the feature maps of the walk kernel
\begin{equation*}
	\varphi_{\text{walk}}^{(k+1)}(u)=\varphi_{\delta}(a(u))\otimes \sum_{v\in\Ncal(u)} \varphi_{\text{walk}}^{(k)}(v),
\end{equation*}
where $\otimes$ denotes the tensor product.

\subsection{Discriminative power between walk kernel and WL subtree kernel}
Before proving the Theorem~\ref{thm:subtree_walk}, let us first show that the WL subtree kernel is always more discriminative than the walk kernel.
\begin{prop}\label{prop:subtree_walk}
 	For any node $u$ in graph $G$ and $u'$ in graph $G'$ and any $k\geq 0$, then $d_{\kappa_{\text{subtree}}^{(k)}}(u,u')=0\implies d_{\kappa_{\text{walk}}^{(k)}}(u,u')=0$.
\end{prop}
This proposition suggests that though both of their feature maps are not injective (see e.g. \citet{kriege2018property}), the feature map of $\kappa_{\text{subtree}}^{(k)}$ is more injective in the sense that for a node $u$, its collision set $\{u'\in \Vcal \,|\, \varphi(u')=\varphi(u) \}$ for $\kappa_{\text{subtree}}^{(k)}$, with $\varphi$ the corresponding feature map, is included in that for $\kappa_{\text{walk}}^{(k)}$. Furthermore, if we denote by $\hat{\kappa}$ the normalized kernel of $\kappa$ such that $\hat{\kappa}(u,u')=\kappa(u,u')/\sqrt{\kappa(u,u)\kappa(u',u')}$, then we have
\begin{corollary}\label{cor:subtree_walk}
 	For any node $u$ in graph $G$ and $u'$ in graph $G'$ and any $k\geq 0$, $d_{\kappa_{\text{subtree}}^{(k)}}(u,u')\geq d_{\hat{\kappa}_{\text{walk}}^{(k)}}(u,u')$.
\end{corollary}
\begin{proof}
	We prove by induction on $k$. It is clear for $k=0$ as both kernels are equal to the Dirac kernel on the node attributes.
	Let us suppose this is true for $k\geq 0$, we will show this is also true for $k+1$.
	We suppose $d_{\kappa_{\text{subtree}}^{(k+1)}}(u,u')=0$. Since $\kappa_{\text{subtree}}^{(k+1)}(u,u)=1$, by equality~\eqref{eq:recursion_wl} we have
	\begin{equation*}
		\begin{aligned}
			1=\kappa_{\text{subtree}}^{(k+1)}(u,u')=\kappa_{\text{subtree}}^{(k)}(u,u')\max\limits_{R\in\Mcal(u,u')} \prod\limits_{(v, v')\in R} \kappa_{\text{subtree}}^{(k)}(v,v'),
		\end{aligned}
	\end{equation*}
	which implies that $\kappa_{\text{subtree}}^{(k)}(u,u')=1$ and $\max_{R\in\Mcal(u,u')} \prod_{(v, v')\in R} \kappa_{\text{subtree}}^{(k)}(v,v')=1$. Then $\delta(a(u), a'(u))=1$ by the non-growth of $\kappa_{\text{subtree}}^{(k)}(u,u')$ on $k$ and it exists an exact matching $R^\star \in\Mcal(u,u')$ such that $|\Ncal(u)|=|\Ncal(u')|=|R^\star|$ and $\forall (v,v')\in R^\star$, $\kappa_{\text{subtree}}^{(k)}(v,v')=1$. Therefore, we have $d_{\kappa_{\text{walk}}^{(k)}}(v,v')=0$ for all $(v,v')\in R^\star$ by the induction hypothesis.
	
	On the other hand, by Lemma~\ref{lm:walk} we have
	\begin{equation*}
		\begin{aligned}
			\kappa_{\text{walk}}^{(k+1)}(u, u')&=\delta(a(u),a'(u')) \sum_{v\in\Ncal(u)} \sum_{v'\in\Ncal(u')} \kappa_{\text{walk}}^{(k)}(v, v') \\
			&=\sum_{v\in\Ncal(u)} \sum_{v'\in\Ncal(u')} \kappa_{\text{walk}}^{(k)}(v, v'),
		\end{aligned}
	\end{equation*}
	which suggest that the feature map of $\kappa_{\text{walk}}^{(k+1)}$ can be written as $\varphi_{\text{walk}}^{(k+1)}(u)=\sum_{v\in\Ncal(u)} \varphi_{\text{walk}}^{(k)}(v)$.
	Then we have 
	\begin{equation*}
		\begin{aligned}
			d_{\kappa_{\text{walk}}^{(k+1)}}(u,u')&=\left\| \sum_{v\in\Ncal(u)} \varphi_{\text{walk}}^{(k)}(v) - \sum_{v'\in\Ncal(u')} \varphi_{\text{walk}}^{(k)}(v') \right\| \\
			&=\left \| \sum_{(v,v')\in R^\star}  \varphi_{\text{walk}}^{(k)}(v) - \varphi_{\text{walk}}^{(k)}(v') \right\| \\
			&\leq \sum_{(v,v')\in R^\star} \| \varphi_{\text{walk}}^{(k)}(v) - \varphi_{\text{walk}}^{(k)}(v') \| \\
			&=\sum_{(v,v')\in R^\star} d_{\kappa_{\text{walk}}^{(k)}}(v,v')=0.
		\end{aligned}
	\end{equation*}
	We conclude that $d_{\kappa_{\text{walk}}^{(k+1)}}(u,u')=0$.
	
Now let us prove the Corollary~\ref{cor:subtree_walk}. The only values that $d_{\kappa_{\text{subtree}}^{(k)}}(u,u')$ can take are 0 and 1. Since $d_{\hat{\kappa}_{\text{walk}}^{(k)}}(u,u')$ is always not larger than 1, we only need to prove $d_{\kappa_{\text{subtree}}^{(k)}}(u,u')=0\implies d_{\hat{\kappa}_{\text{walk}}^{(k)}}(u,u')=0$, which has been shown above.
\end{proof}

\subsection{Proof of Theorem~\ref{thm:subtree_walk}}
Note that using our notation here, $\varphi_1=\varphi_{\text{walk}}^{(k)}$
\begin{proof}
	We prove by induction on $k$. For $k=0$, we have for any $u\in\Vcal$ and $u'\in\Vcal'$
	\begin{equation*}
		\kappa_{\text{subtree}}^{(0)}(u,u')=\delta(a(u),a'(u'))=\delta(\varphi_{\text{walk}}^{(0)}(u), \varphi_{\text{walk}}^{(0)}(u')).
	\end{equation*}
	Assume that \eqref{eq:link_subtree_walk} is true for $k\geq 0$. We want to show this is also true for $k+1$.
	As the only values that the $\delta$ kernel can take is 0 and 1, it suffices to show the equality between $\kappa_{\text{subtree}}^{(k+1)}(u,u')$ and $\delta(\varphi_{\text{walk}}^{(k+1)}(u), \varphi_{\text{walk}}^{(k+1)}(u'))$ in these two situations.
	\begin{itemize}
	\item If $\kappa_{\text{subtree}}^{(k+1)}(u,u')=1$, by Proposition~\ref{prop:subtree_walk} we have $\varphi_{\text{walk}}^{(k+1)}(u)=\varphi_{\text{walk}}^{(k+1)}(u')$, and thus $\delta(\varphi_{\text{walk}}^{(k+1)}(u), \varphi_{\text{walk}}^{(k+1)}(u'))=1$.
	\item If $\kappa_{\text{subtree}}^{(k+1)}(u,u')=0$,
	by the recursive relation of the feature maps in Lemma~\ref{lm:walk}, we have
	\begin{equation*}
		\delta(\varphi_{\text{walk}}^{(k+1)}(u), \varphi_{\text{walk}}^{(k+1)}(u'))=\delta(a(u),a'(u'))\delta\left(\sum_{v\in\Ncal(u)} \varphi_{\text{walk}}^{(k)}(v), \sum_{v'\in\Ncal(u')} \varphi_{\text{walk}}^{(k)}(v')\right).
	\end{equation*}
	By Lemma~\ref{lm:wl_rec}, it suffices to show that
	\begin{equation*}
		\max_{R\in \Mcal(u,u')}\prod_{(v,v')\in R} \kappa_{\text{subtree}}^{(k)}(u,u') =0 \implies \delta\left(\sum_{v\in\Ncal(u)} \varphi_{\text{walk}}^{(k)}(v), \sum_{v'\in\Ncal(u')} \varphi_{\text{walk}}^{(k)}(v')\right)=0.
	\end{equation*}
	The condition $|\Mcal(u,u')|=1$ suggests that there exists exactly one matching of the neighbors of $u$ and $u'$. Let us denote this matching by $R$. The left equality implies that there exists a non-empty subset of neighbor pairs $S\subseteq R$ such that $\kappa_{\text{subtree}}^{(k)}(v,v')=0$ for any $(v,v')\in S$ and $\kappa_{\text{subtree}}^{(k)}(v,v')=1$ for all $(v,v')\notin S$. Then by the induction hypothesis, $\varphi_{\text{walk}}^{(k)}(v)=\varphi_{\text{walk}}^{(k)}(v')$ for all $(v,v')\notin S$ and $\varphi_{\text{walk}}^{(k)}(v)\neq \varphi_{\text{walk}}^{(k)}(v')$ for all $(v,v')\in S$.
	Consequently, $\sum_{(v,v')\notin S}\varphi_{\text{walk}}^{(k)}(v) - \varphi_{\text{walk}}^{(k)}(v')=0$. Now we will show $\sum_{(v,v')\in S}\varphi_{\text{walk}}^{(k)}(v) - \varphi_{\text{walk}}^{(k)}(v')\neq 0$ since all neighbors of either $u$ or $u'$ have distinct attributes. Then
	\begin{equation*}
		\begin{aligned}
			& \| \sum_{v\in\Ncal(u)} \varphi_{\text{walk}}^{(k)}(v) - \sum_{v'\in\Ncal(u')} \varphi_{\text{walk}}^{(k)}(v') \| \\
			=& \| \sum_{(v,v')\in R}  \varphi_{\text{walk}}^{(k)}(v) - \varphi_{\text{walk}}^{(k)}(v') \| \\
			=& \| \sum_{(v,v')\in S}\varphi_{\text{walk}}^{(k)}(v) - \varphi_{\text{walk}}^{(i)}(v') \| >0.
		\end{aligned}
	\end{equation*}
	Therefore, $\delta\left(\sum_{v\in\Ncal(u)} \varphi_{\text{walk}}^{(k)}(v), \sum_{v'\in\Ncal(u')} \varphi_{\text{walk}}^{(k)}(v')\right)=0$.
	\end{itemize}
\end{proof}

\end{document}